\documentclass{article}

\usepackage{arxiv}

\usepackage[utf8]{inputenc} 
\usepackage[T1]{fontenc}    
\usepackage{hyperref}       
\usepackage{url}            
\usepackage{booktabs}       
\usepackage{amsfonts}       
\usepackage{nicefrac}       
\usepackage{microtype}      
\usepackage{lipsum}		
\usepackage{graphicx}
\usepackage{natbib}
\usepackage{doi}

\usepackage{mathtools}
\usepackage{algorithm}
\usepackage{algorithmic}
\usepackage{cleveref}
\usepackage{graphicx}

\usepackage{multirow}
\usepackage{amsthm,amsmath,amsfonts}
\usepackage{caption}
\usepackage{xcolor}
\usepackage{url}
\usepackage[normalem]{ulem}
\usepackage{rotating}
\usepackage{float}
\usepackage{subfigure}
\usepackage{subcaption}

\newtheorem{definition}{Definition}

\title{Behaviour Planning: A Toolkit for Diverse Planning}

\author{Mustafa F. Abdelwahed \\
	School of Computer Science\\
	University of St Andrews\\
	United Kingdom \\
	\texttt{ma342@st-andrews.ac.uk} \\
	\AND
	Joan Espasa \\
    School of Computer Science\\
	University of St Andrews\\
	United Kingdom \\
	\texttt{jea20@st-andrews.ac.uk} \\
    \AND
    Alice Toniolo \\
    School of Computer Science\\
	University of St Andrews\\
	United Kingdom \\
	\texttt{a.toniolo@st-andrews.ac.uk} \\
	\AND
    Ian P. Gent \\
    School of Computer Science\\
	University of St Andrews\\
	United Kingdom \\
	\texttt{ian.gent@st-andrews.ac.uk} 
}

\author{
Mustafa F. Abdelwahed, Joan Espasa, Alice Toniolo, Ian P. Gent \\
School of Computer Science, University of St Andrews, United Kingdom \\
\texttt{\{ma342, jea20, a.toniolo, ian.gent\}@st-andrews.ac.uk}
}

\renewcommand{\shorttitle}{Behaviour Planning}

\hypersetup{
pdftitle={Behaviour Planning: A Toolkit for Diverse Planning},
}

\begin{document}
\maketitle

\begin{abstract}
    Diverse planning approaches are utilised in real-world applications like risk management, automated streamed data analysis, and malware detection. 
    The current diverse planning formulations encode the diversity model as a distance function, which is computational inexpensive when comparing two plans. However, such modelling approach limits what can be encoded as measure of diversity, as well as the ability to explain why two plans are different.
    %
    This paper introduces a novel approach to the diverse planning problem, allowing for more expressive modelling of diversity using a n-dimensional grid representation, where each dimension corresponds to a user-defined feature.
    Furthermore, we present a novel toolkit that generates diverse plans based on such customisable diversity models, called \emph{Behaviour Planning}.
    %
    We provide an implementation for behaviour planning using planning-as-satisfiability. 
    %
    An empirical evaluation of our implementation shows that behaviour planning significantly outperforms the current diverse planning method in generating diverse plans measured on our new customisable diversity models.
    %
    Our implementation is the first diverse planning approach to support planning categories beyond classical planning, such as over-subscription and numerical planning.
\end{abstract}


Automated planners aim to find a sequence of actions that transforms an initial state into a goal state. This sequence of actions is referred to as a plan. The main goal of diverse planners is to generate multiple distinct plans, often called diverse plans, for a given task. Users may require diverse plans for several reasons. One of those reasons is to help account for possible future situations~(\cite{haessler1991cutting}) that might arise. Additionally, side-information such as preferences can be challenging to incorporate into the planning process~(\cite{nguyen2012generating}). By providing a diverse set of options, users can select the plans that best align with their specific preferences and requirements.
From a practical viewpoint, a single optimal solution may be practically challenging to implement. Thus, having a set of different solutions becomes more sensible for the user to pick from~(\cite{ingmar2020modelling,cully2017quality}). Furthermore, many real-world applications drive the need for new diverse planning approaches and planners. Risk management~(\cite{sohrabi2018ai}) exemplifies this, as it requires anticipating various future scenarios and relies on diverse planners to generate multiple potential outcomes.
Another critical application is malware detection~(\cite{boddy2005course,sohrabi2013hypothesis}), where diverse planners are used to identify malicious activity within network streams. In this application, a planner tries to generate a plan matching a given network stream to detect malicious activity. Thus, using a diverse planner helps detect several malicious activities rather than one. 
Several other applications, such as pipeline generation learning in machine learning~(\cite{katz2020exploring}) and business process automation~(\cite{chakraborti2020robotic}), rely on diverse planning for their operations.

There are several diverse planning frameworks to generate $k$ optimal/sub-optimal, diverse plans. A well-established framework biases the planner to search for diverse plans based on a distance function used to model diversity~(\cite{srivastava2007domain,roberts2013tale}). However, using a distance-based diversity model during plan generation does not guarantee the generation of optimal or sub-optimal diverse plans~(\cite{vadlamudi2016combinatorial}). Another framework addresses this issue by adopting a two-stage approach~(\cite{vadlamudi2016combinatorial}). In this approach, the first stage generates a set of plans with a bounded cost, while the second stage extracts a subset of plans based on the provided diversity model. This approach guarantees that all plans are cost-bounded (i.e., optimal/sub-optimal).
Nevertheless, this approach presents several challenges. These include difficulties in modelling diversity, explaining the differences between generated plans, finding plans based on the provided diversity model, and its focus on classical planning. These limitations are primarily concerned with how the current diversity planning model is formulated~(\cite{katz2020reshaping}).
More concretely, diversity is modelled using distance functions, which receive two plans and return a numeric value representing their difference. According to~\citet{coman2011generating}, this modelling approach limits the level of detail that can be expressed. Moreover, such models are incapable of explaining the distinction between different plans, since they fold all the user's aspects of diversity into a single numerical representation and lack the ability to perform reversible computation. Finally, the two-stage diverse planning framework does not incorporate the diversity model when generating plans, which could result in a set of plans that lacks more diverse options. 

In this paper, we address the limitations in previous work by introducing a novel formulation of diversity that enables more expressive modelling. This approach focuses on incorporating user-defined features that represent diverse preferences for plans into the planning generation process. 
By tying diversity to specific, identifiable features in the user's model,  clear explanations for why two plans are different can be easily produced,  essentially making the diversity transparent and interpretable.  
Providing explanations is beneficial for human-in-the-loop systems, enabling users to comprehend the system’s decisions. \citeauthor{sreedharan2020emerging} introduced three user personas for explainable automated planning applications: (i) the end user, who interacts with the planning system through a user interface; (ii) the domain designer, who sets high-level mission objectives for the planning system; and (iii) the algorithm designer, who generates plans based on the end user and domain designer’s requirements. In this work, we are inspired by those personas. Yet, we assume the end user may or may not provide a description on how they would distinguish between plans, and the domain designer is also a domain expert. If the end user provides a diversity description, the domain expert would translate it into a set of features that capture the end user’s definition of diversity, in addition to including their description of diversity, resulting in a single diversity model. Finally, the algorithm designer would integrate these features into the planner to generate diverse plans based on these features.
Therefore, we reformulate the diverse planning problem, replacing the traditional distance threshold with a customisable diversity model, which addresses the limitations in expressivity inherent in the distance function representation, while also offering explanation capabilities.
To this end, we introduce \emph{Behaviour Planning}, a framework that considers the diversity model during planning while guaranteeing optimality, resulting in more diverse plans compared to existing frameworks. 
Our toolkit comprises Behaviour Sorts Suite (\texttt{BSS}) and Forbid Behaviour $\text{Iterative}_\texttt{X}$ ($\texttt{FBI}_\texttt{X}$). \texttt{BSS} is a qualitative-based framework that describes a diversity model while $\texttt{FBI}_\texttt{X}$ is a planning approach that uses the \texttt{BSS}'s diversity model generate diverse plans. 

This paper's contributions are (1) a new formulation for the diverse planning problem, (2) a framework to solve the new problem formulation and (3) an implementation of the framework using planning-as-satisfiability that supports various planning variants such as classical, numeric and over-subscription. We organise this paper as follows: \Cref{sec:motivating-example} motivates the need for the new problem formulation. \Cref{sec:related-work} covers the related work for diverse planning. \Cref{sec:problem-formulation} introduces the new planning problem formulation. \Cref{sec:behaviour-planning} formalises \textit{behaviour planning}. \Cref{sec:behaviour-planning-realisation} presents one way of implementing behaviour planning using planning-as-satisfiability. \Cref{sec:exp-diss} defines the experiments performed, \Cref{sec:discussion} holds the discussion and \Cref{sec:conclusion-future} contains the conclusions and possible future work.

\section{Motivating Example}\label{sec:motivating-example}

\begin{figure}[!hbp]
    \includegraphics[width=\linewidth]{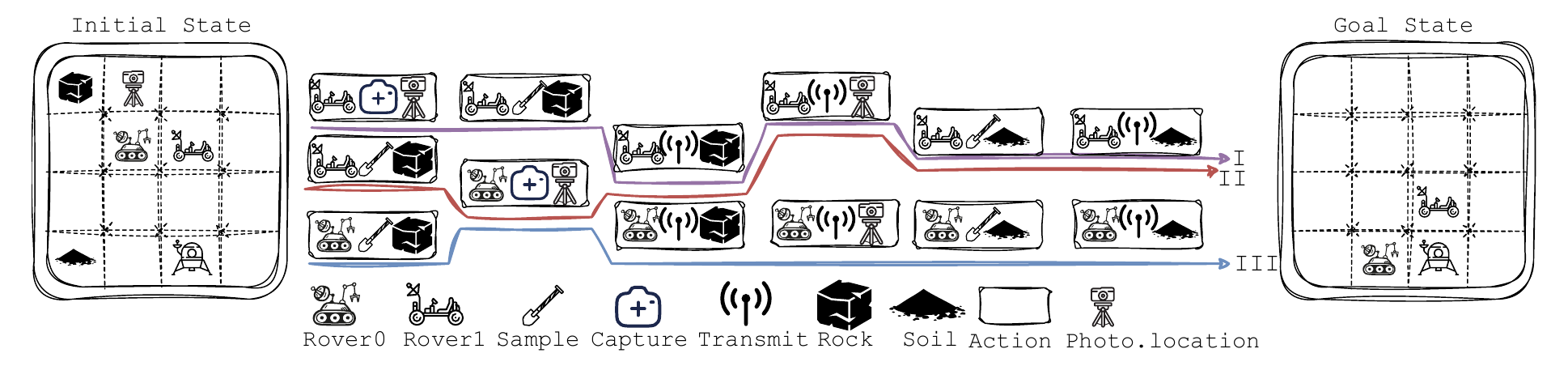}
\caption{Three diverse plans for the Mars rover planning task. Some actions are removed in favour of illustration. See supplementary materials for full plans. Each plan can be read as a sequence of subject-verb-object actions. For example, the first action of plan C is rover1 samples rock, while the last one is rover1 transmits soil. The arrows reflect the temporal sequence of these actions, progressing from the initial state (left) to the goal state (right) where samples and images have been successfully collected and transmitted.}
    \label{fig:d-motivation-example}
\end{figure}

We illustrate our reformulation of the diverse planning problem using the Mars rover planning problem from the third International Planning Competition~(\cite{long20033rd}), where a rover explores Mars and transmits observations to Earth. The goal is to determine a sequence of actions for collecting rock and soil samples, capturing images, and communicating the analysed results to Earth. Assume that a number of plans are required to understand how to best transmit observations back to Earth. Drawing from the aforementioned personas, the recipients of these plans will be the Astronaut on Mars, who will serve as the end user, while the Mission Scientist on Earth will play the role of domain expert. Finally, the planning expert will be the algorithm designer.

\begin{figure}
    \centering
    \includegraphics[scale=1.09]{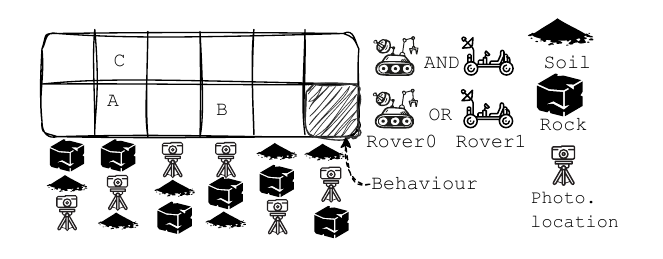}
    \caption{
    A 2D grid showing diversity in the rover domain based on expert-defined dimensions. The horizontal axis represents the order in which samples are transmitted (\texttt{soil}, \texttt{rock}, \texttt{image} from top to bottom). The vertical axis represents the number of rovers used in each plan. The grid positions correspond to the three plans (A, B, and C) shown in \Cref{fig:d-motivation-example}, demonstrating how different planning approaches can be visualised according to these diversity dimensions.
    }
    \label{fig:bspace-rover-example}
\end{figure}

When planning the mission, the astronaut evaluates plans based on different criteria that matter to their objectives. Given their experience on driving rovers, they might want to see all possible transmission orders and select one that suits their needs or preferences. They may prefer transmitting images first, followed by rock and soil data, because image processing takes longer than analysing other samples. Alternatively, some other users might want all data transmitted simultaneously without concern for processing time. The sample collection order could also impact the rover's battery life, and the choice of which rover(s) to use affects the mission budget.
Once the astronaut identifies these important criteria, the mission scientist translates them into measurable dimensions for generating diverse plans. For transmission order, the scientist calculates the possible sequences based on the collected samples. For rover selection, the options could be using Rover0, Rover1, or both rovers together. These two aspects (transmission order and rover choice) become the dimensions that define plan diversity.
The planning expert then incorporates these dimensions into the planner to generate plans that vary according to these features. \Cref{fig:d-motivation-example} shows three example plans for the rover problem, each using different rovers and sample transmission orders. Our key contribution is representing the astronaut's diversity requirements as a discretised grid, where each dimension corresponds to one of these important features.

\Cref{fig:bspace-rover-example} represents the grid for this example on the two dimensions: transmission order and rover number. This grid replaces the distance function used to represent diversity.
Each box within this grid corresponds to a specific property. For instance, plan \texttt{A} employs one of the two rovers and transmits the samples in the following sequence: \texttt{rock}, \texttt{image}, \texttt{soil}.  We call any box in this grid a \emph{behaviour} (i.e., specific property), and the entire grid is called the \emph{behaviour space}. Based on this grid, the astronaut can discern the differences between plans \texttt{A}, \texttt{B}, and \texttt{C} and select the one that aligns with their requirements. For example, the difference between plan \texttt{A} and \texttt{C} would be the number of rovers used to collect the samples, which means that plan \texttt{C} has a higher operating budget than \texttt{A} from the astronaut's perspective.
In \Cref{sec:problem-formulation}, we provide the formulation for the new planning problem based on this behaviour space after covering the related work in \Cref{sec:related-work}. 

\section{Related Work}\label{sec:related-work}
This section delves into the current formulation of the diverse planning problem. It is followed by a literature review that examines the approaches for generating diverse plans and the distance functions employed to model diversity. Additionally, this section includes an overview of other planning variants beyond the classical version.

\subsection{Problem Formulation}

Following \citet{ghallab2016automated},  a planning task is a tuple of $\Xi=\langle S, A, \gamma, \operatorname{cost}, I, G\rangle$, where the first part represents the domain with $S$ as a set of states, $A$ as a set of actions, and $\gamma$ as a transition function $\gamma: S \times A \rightarrow S$ that associates each state $s\in S$ and action $a\in A$ to the next state $\gamma(s,a)=s'$. The function $\operatorname{cost}:  A \rightarrow \mathbb{R}^+$ represents the cost of an action. Note that here we only consider action costs that are independent of the state unlike state-dependent action ones \cite{symk-with-expressive}.  $I\in S$ represents the initial state while $G$ is a formula that models all possible goal states.
A solution for $\Xi$ is a \emph{plan} $\pi$, defined as a sequence of actions $\pi = a_1, a_2, \ldots, a_m$ such that $a_i\in A$ and $\gamma(\gamma(\gamma(I, a_0),\ldots),a_n) \models G$.
A plan's cost is computed by accumulating the actions' costs of $\pi$, such that $\operatorname{cost}(\pi)=\sum_{a\in\pi}cost(a)$. We overload the notation of $\operatorname{cost}$ for simplicity.

The initial version of the diverse planning problem was introduced by \citet{srivastava2007domain}. The authors introduced two versions of the planning problem to allow for maximising or minimising diversity. In this work, we focus on maximising diversity. Assume that $\Pi_\Xi=\{\pi\ |\ \pi\text{ is a solution for }\Xi\}$ represents all plans that solve $\Xi$. \citeauthor{srivastava2007domain} formulated the problem as follows: given a planning problem $\Xi$ and a distance threshold $d$, find a set of plans $\Psi_\Xi\subseteq\Pi_\Xi$ such that the minimum distance between any pairwise plans is greater than a threshold $d$. This problem formulation did not consider the plans' quality. \citet{vadlamudi2016combinatorial} extended the formulation to account for the plan's cost, formulating the problem as: given a planning problem $\Xi$, a distance threshold $d$ and a cost bound $c$, find a set of plans $\Psi_\Xi\subseteq\Pi_\Xi$ such that the minimum distance between any pairwise plans is greater than $d$ and no plan with $\operatorname{cost}(\pi)$ in $\Psi_\Xi$  exceeds the cost $c$. \citet{katz2020reshaping} suggested a taxonomy for this formulation that addresses the satisfying and optimality aspects of the values $c$ and $d$.

\subsection{Frameworks}
The work by \citet{srivastava2007domain} is one of the first attempts to generate diverse plans. 
They proposed generating diverse plans through LPG~(\cite{gerevini2003planning}), a local search-based planner. \citeauthor{srivastava2007domain} used LPG with distance functions to force the planner to generate different plans. Even though they succeeded, the planner converged to non-optimal solutions. \citet{roberts2014evaluating} suggested using a multi-queue $A^\ast$ algorithm to find diverse plans while ensuring optimal results. One queue was for optimality, and the other for diversity. This approach balances the trade-off between diversity and optimality. Furthermore, they favour parsimony (shorter plans) to prevent the domination of one heuristic over the others. 

\citet{vadlamudi2016combinatorial} suggested splitting the problem into two optimisation phases: planning and diversity extraction phases.  In their bi-level optimisation approach, the first phase generates many plans, and the second extracts a subset of diverse plans. This framework received much attention since it guaranteed cost-bounded diverse plan sets. For its first phase, it did not require any specific planner to be used to generate plans. However, using top-k~(\cite{speck2020symbolic}), top-q~(\cite{katz-lee-socs2023}), or diverse~(\cite{katz2020reshaping}) planners is a sensible choice in this phase according to~\citet{bounding-q-diversity}. 
\texttt{SymK}~(\cite{speck2020symbolic}) is a top-k symbolic search planner that groups similar states and systematically explores the search space until it finds a goal state. This grouping facilitates the search for multiple solutions.
$\texttt{K}^\ast$~(\cite{katz-lee-socs2023}) is a top-q planner that efficiently finds and concisely represents a set of all plans with bounded quality for a specified absolute bound.
\texttt{FI}~(\cite{katz2020reshaping}) is a diverse planner that generates multiple plans through a plan-forbid loop. This loop generates a plan and then reformulates the planning task to prompt the planner to find an alternative solution.

To extract a subset of diverse plans from a given set of plans, \citet{katz2020reshaping} suggested a greedy method for extracting a diverse set of plans while maximising the sum of pairwise distances between plans. This value, known as the $\texttt{MaxSum}$ value, quantifies the diversity of a set of plans given a distance function~(\cite{nguyen2012generating}).  \citet{zhong2024bi} proposed a bi-criteria framework for selecting diverse plans in two-stage diverse planning, which is a generalisation of \citet{katz2020reshaping}'s work.

\subsection{Distance Functions}
According to \citet{coman2011generating}, diversity between plans can be measured using both quantitative and qualitative approaches.
The main distinction between the two is that a quantitative metric is not specific to any particular domain and does not need a lot of knowledge engineering. Still, it may not capture differences in plans that matter to domain experts. On the other hand, qualitative plan diversity is based on domain-specific characteristics, making it more practically valuable, but requiring a significant amount of knowledge engineering.

\textbf{Quantitative approaches.} \citet{nguyen2012generating}~proposed measuring the diversity of a set of plans based on the minimal, maximal, or average distance between plans. This distance can be calculated based on the disparity between plans' actions, states, or causal links\footnote{A causal link is a relationship between two actions, where one action establishes a precondition of another action.
}. 
According to \citet{srivastava2007domain}, the distance function can be computed as one minus the Jaccard measurement (i.e., $1-J(\pi_a,\pi_b)$)
\footnote{A Jaccard measurement of two sets is defined as the ratio of the size of their intersection to the size of their union (e.g., $J(a,b)=\vert a\cap b\vert/\vert a \cup b\vert$).} of the plans' actions, states, or causal links. However, there are other suggested distance functions aside from these. 
\citet{roberts2014evaluating} proposed an additional distance function called uniqueness. This function considers two plans to be different if at least one action exists in one plan but not in the other; otherwise, both plans are identical.  
\citet{goldman2015measuring} proposed the Normalised Compression Distance (NCD) to estimate the Kolmogorov complexity and measure the distance between plans. Unlike previous methods, NCD has solid theoretical support. However, a significant drawback is its computational cost and its sensitivity to the compression algorithm used to compute the distance between plans.
Rather than using a distance function to quantify the diversity of a plan set, \citet{nguyen2012generating} suggested using an Integrated Preference Function~(\cite{carlyle2003quantitative}) to quantify the diversity of a plan set based on partially assigned user's preferences. 

\textbf{Qualitative approaches.} \citet{coman2011generating} proposed a qualitative distance metric that includes only the minimal domain-specific content needed to differentiate plans effectively. For instance, they proposed a distance function that would output one of two plans in a game called Wargus, which does not target the same units (i.e., game character groups), and zero if they do target the same units. 
Unfortunately, no commonly accepted technique exists for identifying appropriate domain-specific measures~(\cite{goldman2015measuring}).

\subsection{Planning Variants}\label{subsection:planning-variants}

Current diverse planning approaches primarily focus on classical planning problems. However, other planning categories would greatly benefit from diversity in planning approaches. In this subsection, we explore two planning variants that would benefit from diverse planning.  

Over-subscription planning (OSP) aims to find plans with maximum value for a given budget or resource~(\cite{smith2004choosing}). This planning variant is essential in real-world situations where resources are limited. 
\citet{katz2019oversubscription} establishes a foundation for adapting heuristics from classical planning to over-subscription planning. They proposed reformulating the over-subscription planning task as a classical planning task with two cost functions for operators. The first cost function represents the original operator costs, ensuring that solutions remain within the specified budget, while the second cost function encodes the goal predicates' utilities
of the problem. \citet{katz2019oversubscription} extension to \texttt{SymK} is considered the first attempt to generate a set of plans for this kind of planning. 
Regarding the over-subscription utility function, \citet{katz2019oversubscription} defines the plan's utility as the sum of achieved goal predicates at the end of the plan. The utility function $u:\Pi_\Xi\rightarrow \mathbb{R}^{+}$ is defined as a function that receives a plan $\pi$ and returns the sum of the utilities at the end of $\pi$. 

Planning with resources and numeric state variables is crucial for efficiently encoding domains involving distances, costs, fuel, and other quantitative aspects. Such information is represented by a numeric fluent function that maps a collection of objects to a rational number~(\cite{fox2003pddl2}). Numeric fluents are a helpful tool for modelling the quantitative properties of objects.
The IPC2023 competition showed progress with numerical planning~(\cite{taitler20242023}), yet as far as we know, no diverse or top-k planners support numerical planning.

\section{A novel diverse planning problem formulation}\label{sec:problem-formulation}

This section proposes our novel diversity planning framework which accommodates a more expressive diversity modelling approach.   

\citet{lehman2011abandoning} collapsed the search space for combinatorial optimisation problems into a finite space (\emph{behaviour space}) to model solutions' characteristics. 
Their motivation was to keep track of different solutions generated by evolutionary algorithms.
They used behaviour spaces to include or discard solutions into an archive based on their objective values in case the behaviour was already included. Inspired by such ideas,  our key novelty is to represent  diversity using an \emph{n-dimensional grid} as illustrated in the motivating example. We refer to this grid as the \emph{behaviour space}, and each box within the grid is called a \emph{behaviour}. We use behaviour space as a guide for the planner to search for plans that satisfy a desired property, guaranteeing the generation of a diverse plan set that contains plans with different behaviours.

Assume that we have a set of $n$ dimensions $\Delta_\Xi$ relevant to a planning task $\Xi$ where each dimension $\Delta_i\in\Delta_\Xi$ is set of possible values  $\delta^i$. We assume $\Delta_i=\{\delta^i_1,\delta^i_2  \ldots \}$. 
Each dimension $\Delta_i$ corresponds to values that a plan can exhibit based on a user-defined feature deemed important in determining diversity between plans. Behaviours, behaviour  space and plan behaviour are then represented as follows: 

\begin{definition}[Behaviour]\label{def:behaviour}
Given a  planning task $\Xi$ and diversity dimensions $\Delta_\Xi$ where $|\Delta_\Xi|=n$ we define a \emph{behaviour} $\mathcal{B}=\langle\delta^1,\ldots,\delta^n\rangle$ as an $n$-tuple containing a value $\delta^i\in\Delta_i$ for each dimension $\Delta_i\in\Delta_\Xi$. 
\end{definition}

\begin{definition}[Behaviour space]\label{def:bspace}
Given a planning task $\Xi$ and diversity dimensions $\Delta_\Xi$, we define a \emph{behaviour space} $BS_{\Delta_\Xi}$ as the set of all possible combinations (i.e., behaviours) for $\Xi$ where $BS_{\Delta_\Xi} = \Delta_1 \times \Delta_2 \times \ldots \times \Delta_n$. This is equivalent to $\{\mathcal{B}| \mathcal{B} \text{ is a behaviour for $\Delta_\Xi$}\}$. Instead of $BS_{\Delta_\Xi}$, we write $BS_{\Delta}$ for simplicity, where the planning task $\Xi$ is clear from context.
\end{definition}

To extract a plan $\pi$'s behaviour $\mathcal{B}$ for a given behaviour space $BS_\Delta$, we must extract the dimensions' values from $\pi$ and construct its behaviour. Thus, for each dimension we use an extracting function $\odot_i:\Pi_\Xi\rightarrow\Delta_i$ that extracts the value $\delta^i$ of dimension $\Delta_i$ for plan $\pi$. The set $\odot_\Delta$ is a collection of all extracting functions. We can then define a plan's behaviour as follows:


\begin{definition}[Plan behaviour]\label{def:plan-beh}
Given a diverse planning task $\Xi$, $\pi\in \Pi_\Xi$ a valid plan for $\Xi$, and a behaviour space $BS_\Delta$ with diversity dimensions $\Delta_\Xi$ and corresponding extracting functions $\odot_\Delta$, \emph{plan behaviour} is a function returning the behaviour corresponding to the plan. We write this as $\operatorname{PBehaviour}:\odot_\Delta\times\Pi_\Xi\rightarrow BS_\Delta$, where  $\operatorname{PBehaviour}(\odot_\Delta,\pi) = \langle\odot_1(\pi),\ldots,\odot_n(\pi)\rangle$.
\end{definition}

In the context of the motivating example presented in \Cref{sec:motivating-example}, the behaviour space is a 2-dimensional grid, formed by $\Delta_1=\{ \texttt{R-S-I},\texttt{S-I-R},\ldots\}$ representing the order of the collected samples and $\Delta_2=\{1,2,\ldots\}$ indicating the number of resources used. The behaviour space  $BS_\Delta=\{\langle\texttt{R-S-I,1}\rangle, \langle\texttt{R-S-I,2}\rangle,\ldots,\langle\texttt{S-I-R,1}\rangle\}$ encompasses all possible behaviours. Regarding a plan’s behaviour, plan C ($\pi_\texttt{C}$) for example would be represented as $\operatorname{PBehaviour}(\odot_\Delta,\pi_{\texttt{C}})=\langle\texttt{R-I-S}, 2\rangle$.

To quantify the diversity of a plan set $\Psi_\Xi\subseteq\Pi_\Xi$ based on a behaviour space $BS_\Delta$, we introduce a metric called the \emph{behaviour count}. To compute this metric, we first infer the behaviour for each plan $\pi$ on a given set of plans $\Psi_\Xi$ using~\Cref{def:plan-beh}. Then, we count the number of distinct behaviours available in $\Psi_\Xi$. We define  Behaviour Count as follows: 

\begin{definition}[Behaviour count]\label{def:bc}
Given a planning problem $\Xi$, $\Psi_\Xi\subseteq\Pi_\Xi$ as a valid plan set for $\Xi$ and diversity dimensions $\Delta_\Xi$  with extracting functions $\odot_\Delta$, \emph{behaviour count} is defined as $\operatorname{BC}:\odot_\Delta\times\Pi_\Xi\rightarrow\mathbb{N}$, a function that computes the number of available behaviours in $\Psi_\Xi$ calculated as $\operatorname{BC}(\odot_\Delta, \Psi_\Xi) = \left| \{ \operatorname{PBehaviour}(\odot_\Delta, \pi) \mid \pi \in \Psi_\Xi \} \right|$
\end{definition}

After describing how the diversity model is represented using behaviour space, we define the planning problem as follows:

\begin{definition}[diversity planning problem]\label{def:planning-task}
    Given a planning problem $\Xi$, a cost bound $c$, number of plans $k$ and a set of dimensions $\Delta_\Xi$, find a set of plans $\Psi_\Xi\subseteq\Pi_\Xi$ that is subject to $cost(\pi) \leq c\ \forall \pi\in\Psi$, $|\Psi_\Xi|\leq k$ and the behaviour count $BC(\odot_\Delta,\Psi_\Xi)$ of $\Psi_\Xi$ should be maximised. 
\end{definition}
We refer to this diversity planning problem as a planning problem $\Xi$ for simplicity. 

\section{Behaviour Planning} \label{sec:behaviour-planning}
Behaviour planning generates diverse plans based on the provided behaviour space $BS_\Delta$ for the planning problem $\Xi$. It comprises the Behaviour Sorts Suite (\texttt{BSS}) and Forbid Behaviour $\text{Iterative}_\texttt{X}$ ($\texttt{FBI}_\texttt{X}$). \texttt{BSS} represents a behaviour space, while $\texttt{FBI}_\texttt{X}$ uses this model to generate diverse plans.

\subsection{Behaviour Sorts Suite (\texttt{BSS})}
Here, we describe the Behaviour Sorts Suite's components. The first component is behaviour spaces, which allow domain experts to represent a behaviour space based on features of interest. The second component is a library of behaviour features gathered from the literature. 

\subsubsection{Behaviour Spaces from user-defined features}\label{sec:behaviour-spaces}

\texttt{BSS} shows how a plan's behaviour is represented from the features being defined by users. First, the user has to decide how to differentiate between plans based on features. Then, each feature is used to construct a dimension in the behaviour space. The values for each feature should then be discretised. For example, a domain expert would differentiate between plans based on their cost and consider plans with costs between 10 and 20 to be the same. $\Delta_i$ values for such a domain could be integer values where each value represents a range (e.g., value 1 is the continuous value between 10 and 20). In this case, the domain expert decides the granularity of those values.
Now assume a set of features $F_\Xi=\{f_1,\ldots,f_n\}$, where a $f_i$ is a tuple containing $\Delta_i$ and $\odot_i:\Pi_\Xi\rightarrow\Delta_i$ (i.e., $f_i=\langle\Delta_i,\odot_i\rangle$). The $\Delta_i$ represents all possible values for feature $f_i$, and the $\odot_i$ is an extracting function that extracts the value $\delta^i$ of feature $f_i$ from a given plan $\pi$, as described in \Cref{sec:problem-formulation}. Note that additional information might be needed, but is embodied in the extracting function implementation. We refer to this additional information for a feature $f_i$ as $\textit{AddInfo}_i$. An example of additional information could be the maximum cost a plan could achieve. 

\subsubsection{Behaviour Features Library}\label{sec:features-library}
The literature showed attempts to differentiate between plans using domain-independent features. Our features library is an initial collection of some of those features and can be easily extended in the future. 

For classical planning tasks, \citet{mantik2022preference} suggested two primary features: (i) plan length ($f_{cb}$) and (ii) resource utilisation ($f_{ru}$). The plan length feature distinguishes between plans based on the number of actions in each plan. On the other hand, the resource utilisation feature computes the number of used resources in a plan and then uses this information to differentiate between plans. Note that the user provides information on the resources used. \citet{mabdelwahed-pair-2023} suggested differentiating between plans based on the order of the goal predicates achieved by each plan ($f_{go}$). This feature ignores the order of the goal predicates achieved simultaneously.

No features are available to differentiate between plans in the existing literature for over-subscription and numerical planning tasks. Hence, we propose differentiating plans based on their utility values ($f_{uv}$) for over-subscription while utilising numeric fluents defined in the domain model ($f_{nf}$) for numeric ones. In this paper, we consider the same definition provided by \citet{katz2019oversubscription}, and we use the plan's utility as a feature to distinguish between plans.
As for numerical planning tasks, we suggest comparing plans based on  numeric fluents defined in the planning task $\Xi$. Domain experts can select one or more of those fluents to represent the end user's diversity definition. For this feature, we use the values of the provided fluents at the end of the plan. 
In \Cref{sec:behaviour-planning-realisation}, we provide examples implementing and using those features.

\subsection{Forbid Behaviour \texorpdfstring{$\text{Iterative}_\texttt{X}$}{} (\texorpdfstring{$\texttt{FBI}_\texttt{X}$}{})}\label{sec:FBI-section}

Forbid Behaviour $\text{Iterative}_\text{X}$ is a planning approach that uses a behaviour space to generate diverse plans by finding a plan per behaviour. It is inspired by the Forbid Iterative (\texttt{FI}) approach suggested by \citet{katz2018novel}. However, the significant difference between the $\texttt{FBI}_\texttt{X}$ planner and \texttt{FI} is that the $\texttt{FBI}_\texttt{X}$ forbids behaviours instead of plans. Since $\texttt{FBI}_\texttt{X}$ is a generic approach, we use the $\texttt{X}$ symbol to denote the technology utilised for implementing $\texttt{FBI}$.

\begin{figure}
\centering
\subfigure[FI operation]{\label{fig:fi-operation}\includegraphics[width=0.48\linewidth]{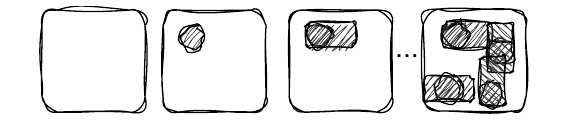}}
\subfigure[FBI operation]{\label{fig:fbi-operation}\includegraphics[width=0.48\linewidth]{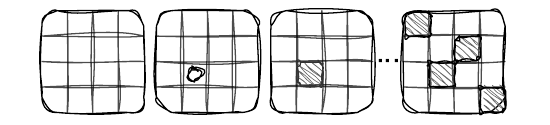}}
\caption{The white space represents all possible plans for a planning task $\Xi$. A black circle represents a plan, while a black box denotes a set of plans. For example, the black box for \texttt{FI} represents all possible reorderings of a plan. On the other hand, the black box for $\texttt{FBI}_\texttt{X}$ represents a set of plans that share the same behaviour.
}

\end{figure}

Figures \ref{fig:fi-operation}-\ref{fig:fbi-operation} illustrate how each planner explores the solution space when solving a planning problem where the goal is to find four plans. \texttt{FI} starts with finding one plan and then forbids it together with its possible reorderings. Then, it keeps repeating this operation until it reaches the number of required plans (\Cref{fig:fi-operation}). On the other hand, $\texttt{FBI}_\texttt{X}$ finds a plan, then uses behaviour space to infer this plan's behaviour and to acquire another plan, it forbids this behaviour, then uses the updated formulation to obtain a different one (\Cref{fig:fbi-operation}). When $\texttt{FBI}_\texttt{X}$  prohibits a particular behaviour, all plans associated with this behaviour, along with their symmetrical and reordered versions that result in the same behaviour, are also forbidden.

Our implementation of $\texttt{FBI}_\texttt{X}$ uses a function called $\operatorname{BehaviourGenerator}$.
This function receives a planning problem $\Xi$, a set of diversity features $F_\Xi$, a set of plans with different behaviours $\Psi_\Xi$ and returns a plan with a behaviour $\mathcal{B} \not \in \{\operatorname{PBehaviour}(\odot_\Delta,\pi)\vert \pi \in \Psi_\Xi\}$ or an empty sequence of actions in case it failed to find a new behaviour ($\pi=\emptyset$). In  \Cref{sec:fbireal}, we shall show how to realise such a function. 
\Cref{alg:fbi-planner-main} describes $\texttt{FBI}_\texttt{X}$'s key operation. It starts with an empty set of plans ($\Psi_\Xi$)  (Line \ref{alg-line:bspace-planner-initial-set}), then it keeps generating and accumulating plans with new behaviours until $\operatorname{BehaviourGenerator}$ returns no plan or $k$ has been reached (Lines \ref{alg-line:fbi-main-loop-start}-\ref{alg-line:fbi-main-loop-end}).

\begin{algorithm}
\caption{$\operatorname{FBI}_\texttt{X}$}\label{alg:fbi-planner-main} 
\begin{algorithmic}[1]
\REQUIRE $\Xi$: Planning task, $F_\Xi$: Diversity features, $k$: Required plans count
\ENSURE $\Psi_\Xi$: Set of plans with different behaviours, $BC$: Behaviour count. 
\STATE $\Psi_\Xi \gets \emptyset; BC\gets 0$\label{alg-line:bspace-planner-initial-set}
\STATE $\textbf{do}$ \label{alg-line:fbi-main-loop-start}
\STATE $\quad \pi \gets \operatorname{BehaviourGenerator}(\Xi, F_\Xi, \Psi_\Xi)$ 
\STATE $\quad \Psi_\Xi \gets \Psi_\Xi \cup \{\pi\}$ 
\STATE $\quad BC \gets BC+1$ 
\STATE $\textbf{while }|\Psi_\Xi|< k \textbf{ and } \pi \not = \emptyset$ \label{alg-line:fbi-main-loop-end}
\RETURN $\Psi_\Xi, BC$
\end{algorithmic}
\end{algorithm}

There could be a scenario where a domain expert requires a specific number of plans that exceed the number of possible behaviours. To resolve such a situation, we use another function to generate plans for a given set of behaviours. Let $\operatorname{PlanGenerator}$
be a function that receives a planning problem $\Xi$ and a set of plans $\Psi_\Xi$, and returns a plan $\pi \not\in \Psi_\Xi$ if one exists not belonging to $\Psi_\Xi$, where $\pi$ is allowed to have a previously generated behaviour. 
\Cref{alg:fbi-k-planner-main} can generate more plans if the required number of plans is bigger than the number of available behaviours by forbidding plans (i.e., preventing the planner from generating the same plans) rather than behaviours. \Cref{alg:fbi-k-planner-main} invokes \Cref{alg:fbi-planner-main} to generate a plan per behaviour and collects those plans behaviours (Line $\ref{alg:fbi-k-get-behaviours}$). Afterwards, $\operatorname{FBI-k}_\texttt{X}$ resumes the same loop as $\texttt{FBI}_\texttt{X}$ except for generating plans rather than behaviours (Lines $\ref{alg:fbi-k-start-loop}$-$\ref{alg:fbi-k-end-loop}$). 

When the required number of plans exceeds the available behaviours (i.e., $k>\vert BS_\Delta\vert$), $\operatorname{FBI-k}_\texttt{X}$ will keep generating plans without accounting for behaviours. Improving upon this aspect of exploration is left for future work. 
 
\begin{algorithm}
\caption{$\operatorname{FBI-k}_\texttt{X}$}\label{alg:fbi-k-planner-main}
\begin{algorithmic}[1]
\REQUIRE $\Xi$: Planning task, $F_\Xi$: Diversity Features, $k$: Required plans count
\ENSURE $\Psi_\Xi$: Set of plans with different behaviours, $BC$: Behaviour count.
\STATE $\Psi_\Xi, BC \gets \operatorname{FBI}_\texttt{X}(\Xi,F_\Xi,k)$  \hfill[where $|\Psi_\Xi|=|BS_\Delta|<k$]\label{alg:fbi-k-get-behaviours}\label{line:using-fbi-first} 
\STATE $\textbf{do}$ \label{alg:fbi-k-start-loop}
\STATE  $\quad\pi\gets \operatorname{PlanGenerator}(\Xi,\Psi_\Xi)$
\STATE $\quad\Psi_\Xi \gets \Psi_\Xi \cup \{\pi\}$
\STATE $\textbf{while }|\Psi_\Xi| < k \textbf{ and } \pi \not= \emptyset$ \label{alg:fbi-k-end-loop}
\RETURN $\Psi_\Xi, BC$
\end{algorithmic}
\end{algorithm}

\section{Realisation of Behaviour Planning}
\label{sec:behaviour-planning-realisation}

We used a planning-as-satisfiability to implement $\operatorname{BehaviourGenerator}$ and $\operatorname{PBehaviour}$. This approach allows us to incorporate arbitrary constraints into the problem, granting us the flexibility to describe and reason over plan behaviours during search.
The first works that addressed the planning-as-satisfiability (model finding) problem~(\cite{kautz1996encoding}) showed that off-the-shelf SAT solvers could effectively solve planning problems. In the last decade, various works followed by leveraging SMT~(\cite{bofill2016rantanplan,leofante2020optimal}), SAT~(\cite{Rintanen12,holler2022encoding}) or CP~(\cite{aiplanningcp,villaret2021exploring}) solvers, among others. 
The problem $\Xi$ is generally solved by considering a sequence of queries in the form of satisfaction problems $\phi_0$, $\phi_1$, $\phi_2$, \ldots, where $\phi_i$ encodes the existence of a plan that reaches a goal state from the initial state in exactly $i$ steps. 
The number of actions allowed per step would vary based on the encoding used. A simple linear encoding allows exactly one action per step, unlike other encodings such as Relaxed $\exists$-step~(\cite{bofill2017relaxed}), which allows more than one action per step. 
 
The solving procedure will sequentially test the satisfiability of $\phi_0$, $\phi_1$, $\phi_2$, \ldots, $\phi_n$, until a satisfiable formula $\phi_n$ is found, proving the existence of a valid plan of exactly $n$ steps. Each formula $\phi_n$ is defined as 

\begin{equation}
\label{equ:planning-as-smt-equ}
\phi_n \coloneqq \mathcal{I}(s_0)\bigwedge_{i=0}^{n-1}\mathcal{T}(s_i,s_{i+1})\wedge
\bigvee_{k=1}^{n}\mathcal{G}(s_k)
\end{equation}
where $\mathcal{I}$ and $\mathcal{G}$ respectively encode the formulas for the initial $I$ and goal $G$ states, and $\mathcal{T}$ encodes the transition function $\gamma$ in terms of the   preconditions ($\operatorname{a}_i\implies \operatorname{pre}_i$) and effects ($\operatorname{a}_i\implies \operatorname{eff}_i$) for each action  $a_i\in A$, and the related frame axioms.
For further details on planning-as-satisfiability, readers are encouraged to refer to~\citet{planningassat}.

In our framework, to identify a plan, we abstract the creation of the formula $\phi_n$ by defining the function $\operatorname{Encoder}(\Xi,n)=\phi_n$, which takes a planning task $\Xi$, a number of steps ($n$) and returns a formula that encodes the existence of a plan with exactly $n$ steps. In this work, we implement the $\operatorname{Encoder}$ function adapting the sequential encoding suggested by~\citet{kautz1992planning}.  

\subsection{Behaviour Space Dimensions Realisation}
Now we describe an example of behaviour space based on a planning-as-SMT approach and constructed over the features presented in \Cref{sec:features-library}: cost bound ($f_{cb}$), resource utilisation ($f_{ru}$), goal predicate ordering ($f_{go}$), utility value ($f_{uv}$), and numeric fluent ($f_{nf}$). Each feature $f_i$ is characterised by its domain $\Delta_i$ and the extracting function $\odot_{i}$ used to obtain a value $\delta^i$. Every dimension has its SMT expression that includes a variable that holds the value $\delta^i$. Thus, we extend $f_i$ to include $\textit{Expr}_i$ to represent the SMT variable containing $\delta^i$, i.e., $f_i=\langle \Delta_i,\odot_i, \textit{Expr}_i\rangle$.

 The planning expert determines which dimensions to use to represent their diversity model. Using planning-as-satisfiability, we can easily append the selected dimensions encodings to the formula $\phi_n$. 
 This allows them to consider the diversity model when generating plans. To retrieve the values from those dimensions, we use $\operatorname{ExtractModel}(\phi_n)=\mathcal{M}_{\phi}$, a function that receives a formula $\phi_n$ and returns a model $\mathcal{M}_{\phi}$. That is, an assignment for all variables in the formula. We use the $[\cdot]$ notation to access the value of a variable from $\phi$ (e.g., $\mathcal{M}_\phi[var]$ denotes the value of variable $var$ in the model $\mathcal{M}_\phi$).
Below, we present the features' descriptions and corresponding information to construct their dimensions and encodings.  

\paragraph{\textbf{Feature I - Cost Bound ($f_{cb}$).}} In this work, we only consider makespan-optimality. Therefore, our cost function $\operatorname{cost}(\pi)$ will encode the length of the plan with no additional information required, except for the cost bound in $\Xi$ (e.g., $c$).  The corresponding cost-bound dimension $\Delta_{cb}$ includes $\Delta_{cb}=\{0,\ldots,n\}$, a set of integer values between 0 and the formula's length ($n$). The cost bound $c$ is less than or equal to n. The extraction function $\odot_{cb}$ returns the number of actions in $\pi$. For this dimension, the $\textit{Expr}_{cb}$ would be $cvalue$.

To compute the cost of a given plan, we use $\omega(i)=v,v\in \{0,1\}$ a function that returns 1 if an action is selected at step $i$ otherwise, it returns 0.
 
$$\Gamma \coloneq  (cvalue = \sum_{i=0}^{n} \omega(i)) \wedge (cvalue \leq c)$$
   
\paragraph{\textbf{Feature II - Resource utilisation ($f_{ru}$).}} This feature requires domain-specific information. We consider the set of problem objects $\textit{AddInfo}_{ru}$, representing the resources of interest. Following the motivating example in \Cref{sec:motivating-example}, $\textit{AddInfo}_{ru}$ would be a set of available rovers in the planning task as shown in \Cref{fig:bspace-rover-example} (e.g. $\textit{AddInfo}_{ru}{=}\{\texttt{Rover0},\texttt{Rover1}\}$). The resource utilisation dimension $\Delta_{ru}$ is formed by $\Delta_{ru}{=}\{0,\ldots,|\textit{AddInfo}_{ru}|\}$, a set of integers between 0 and the number of resources provided in $\textit{AddInfo}_{ru}$. As for the extract function $\odot_{ru}$, it uses the provided information in $\textit{AddInfo}_{ru}$ to count how many resources are used in a given plan $\pi$. This is done by counting the number of actions in $\pi$ that have the resource as a parameter and then returning how many resources are used in $\pi$. The $\textit{Expr}_{ru}$ would be $ru$.

We encode the number of used resources in a plan $\pi$ using an indicator function $\iota(r,i)=v,v\in\{0,1\}$, returning 1 if the resource $r\in \textit{AddInfo}_{ru}$ is used at step $i$, and 0 otherwise. The resource utilisation feature then sums 1 for each $r\in \textit{AddInfo}_{ru}$ if $r$ is used in any plan step. 
 
$$
\Lambda \coloneq ru = \sum_{r\in \textit{AddInfo}_{ru}} \begin{cases} 
    0 & \text{if } \sum_{i=0}^{n} \iota (r, i) = 0 \\
    1 & \text{otherwise }
\end{cases}
$$

Notice in particular how we model the resource utilisation impacts the values of $\Delta_{ru}$. For example, if instead we wanted to compute the ratio of the number of actions using a resource $r$ to the plan length, then $\Delta_{ru}$ would be a set of rational numbers rather than a set of integers.
The information about the available ranges would be provided in $\textit{AddInfo}_{ru}$ alongside the resource information. However, there is no widely accepted approach for modelling resource utilisation for a plan. In this work, we adhere to \citet{mantik2022preference}'s suggestion to distinguish between plans based on the used resources. 

\paragraph{\textbf{Feature III - Goal predicate ordering ($f_{go}$).}} 
This feature considers the total order in which goal predicates are achieved. The goal ordering dimension $\Delta_{go}$ is a set containing all permutations for goal predicates in $G$. The extract function $\odot_{go}$ returns the total ordering of the goal predicates.
To encode such a dimension, let $\operatorname{PStep}(p)=v, v\in\mathbb{N}$ be a function that maps a predicate $p$ to the step $i$ at which it first became true. For a given goal predicate $g\in G$ and step $i$, we can encode its semantics using the formula 
$$ (\operatorname{PStep}(g)=i) \iff  (g_i \bigwedge_{j=0}^{i-1}\neg{g_j})$$
That is, $\operatorname{PStep}(g)=i$ when the goal predicate $g$ is satisfied at step $i$ and not satisfied in any step from $0$ up to $i-1$. In some cases, $g$ can become false after $i$ again, but we are not concerned about that. To support the over-subscription planning for this dimension, the $\operatorname{PStep}(g)$ is set to $-1$ in cases $g$ is not achieved. That is, $(\operatorname{PStep}(g)=-1) \iff  \left( \bigwedge_{j=1}^{n} \neg{g_j} \right), \forall g \in G$.

The most straightforward encoding for this dimension is to create a Boolean variable for each possible ordering, thus resulting in $\vert G \vert !$ Boolean variables. However, such encoding does not scale well. Therefore, we encode a precedence relation between any two goal predicates $g_a, g_b \in G$ by stating $to_{ab} \iff \operatorname{PStep}(g_a) \leq \operatorname{PStep}(g_b)$. The fresh Boolean variable $to_{ab}$ indicates whether $g_a$ is achieved before $g_b$ or not. The conjunction $\bigwedge (to_{ij} =\mathcal{M}_{\phi_n^\prime}[to_{ij}]), \forall g_i,g_j\in G$ encodes a total order between those predicates, and is the $\textit{Expr}_{go}$ variable for this dimension. 
The full encoding of the goal ordering feature $\Upsilon$ is then the conjunction of the three following equations:

\begin{align*}
\Upsilon \coloneq go \iff \bigwedge \left\{
\begin{aligned}
& (\operatorname{PStep}(g)=i) \iff  \left( g_i \bigwedge_{j=0}^{i-1} \neg{g_j} \right) && \forall i \in 1..n, \forall g \in G\\
& (\operatorname{PStep}(g)=-1) \iff  \left( \bigwedge_{j=1}^{n} \neg{g_j} \right) &&  \forall g \in G \\
& to_{ij} \iff (\operatorname{PStep}(g_i) \leq \operatorname{PStep}(g_j)) && \forall g_i, g_j \in G
\end{aligned}
\right.
\end{align*}

\paragraph{\textbf{Feature IV - Utility value ($f_{uv}$).}}
This feature distinguishes between any two given plans based on their utility values. The utility value of a plan $\pi$ is computed using the function $\operatorname{u}:\Pi_\Xi\rightarrow\mathbb{R}^+$ as discussed in \Cref{subsection:planning-variants}. To construct $\Delta_{uv}$, we use the function $\operatorname{uvalue}_\Xi: G\rightarrow \mathbb{R}^+$, which extracts the utility value for a goal predicate $g \in G$ in a planning task $\Xi$.
The corresponding utility value dimension $\Delta_{uv}= \left\{ \sum_{g\in H}  \operatorname{uvalue}_\Xi(g) \vert H \subseteq G \right\}$, a set of all possible utility values reflecting the achieved goal predicates. $\operatorname{Achieve}(g)=v,v\in\{0,1\}$ is a function that receives a goal predicate $g$ and returns 1 if $g$ is achieved and 0 otherwise. 
The extract function $\odot_{uv}$ returns the accumulated sum for goal predicates utilities at the last step of $\pi$. As for the $\textit{Expr}_{uv}$, it would be $uv$.
We encode $\operatorname{uvalue}_\Xi$ to return the utility of a goal predicate $g$, and $\operatorname{Achieve}(g)$ would return 1 if $g$ is achieved at the last step of the formula. Otherwise, it returns 0. We encode this dimension as follows:

$$\Omega\coloneq uv = \sum_{g \in G } \operatorname{Achieve}(g) \cdot \operatorname{uvalue}_\Xi(g) $$%

\paragraph{\textbf{Feature V - Numeric Fluent ($f_\mathit{nf}$).}} 
Since a given planning problem may involve multiple numeric fluents, domain experts may choose one or more to differentiate between plans. Numeric fluents are functions that maintain a value for a variable. This feature  explains how a  numeric fluent is discretised and represents a type as it can be used for any numeric fluent.    
This feature is similar to $f_{ru}$ in requiring domain-specific information specifying which fluent to use when extracting a plan's behaviour. This information is passed in $ \textit{AddInfo}_{\mathit{nf}}^{\texttt{var}}= \langle \texttt{var}, \operatorname{min}, \operatorname{max}, \varepsilon \rangle$ as a tuple containing the fluent variable ($\texttt{var}$  appearing in a state $s\in S$), the minimum value ($\operatorname{min}\in \mathbb{R}$), maximum value ($\operatorname{max}\in \mathbb{R}$) and an incremental value ($\varepsilon\in \mathbb{R}$). To encode all possible ranges for a given variable $\operatorname{var}$, we use an integer variable called ${box}_{\texttt{var}}$. This integer variable represents a point in the dimension $\Delta_\mathit{nf}^{\texttt{var}}$ and takes values  $\Delta_\mathit{nf}^{\texttt{var}} = \left\{ \operatorname{min} + i \cdot \varepsilon \;\middle|\; i \in \mathbb{N},\; \operatorname{min} + i \cdot \varepsilon \leq \operatorname{max} \right\}$. Each value represents a range. Note that each fluent $\operatorname{var}$ has its own ${box}_{\texttt{var}}$. The function $\odot_{\mathit{nf}}^{\texttt{var}}$ returns the ${box}_{\texttt{var}}$ value at the end of the plan and the $\textit{Expr}_\textit{nf}$ is ${box}_{\texttt{var}}$. We encode this discretisation as follows:

$$
\Sigma_{\texttt{var}} \coloneq \bigwedge_{i = 0}^{(\operatorname{max}/\varepsilon)-1} \left( \operatorname{var} \geq \operatorname{min} + i \cdot \varepsilon \;\land\; \operatorname{var} < \operatorname{min} + (i+1)\cdot \varepsilon \;\Rightarrow\; ({box}_{\texttt{var}} = i) \right)
$$

To clarify this, assume $\textit{AddInfo}_\mathit{\texttt{foo}}=\langle\texttt{foo}, 0, 10, 2 \rangle$. The possible values of ${box}_{\texttt{foo}}$ are $\Delta_\mathit{nf}^{\texttt{foo}}=\{0,1,2,3,4\}$ where ${box}_{\texttt{foo}}=0$ indicates that $0 \leq \texttt{foo} < 2$, ${box}_{\texttt{foo}}=1$ indicates that $2 \leq \texttt{foo} < 4$, and so on until ${box}_{\texttt{foo}}=4$ indicates that $8 \leq \texttt{foo} \leq 10$. Note that the maximum value should be included; that is why the maximum value of ${box}_{\texttt{foo}}$ is less than or equal to the value $\operatorname{max}$ defined in $\textit{AddInfo}_\mathit{\texttt{foo}}$. 

\paragraph{\textbf{Encoding Optimisations.}} 
To reduce the impact of the dimensions encoding on the planner's performance, we suggest reducing the number of non-Boolean variables as much as possible. For instance, encoding $\Gamma$ as a sum of all enabled actions in each step creates an overhead that significantly affects the coverage. We use an auxiliary ($\omega$) function to return 1 if any action is selected for a given step; otherwise, we return 0. To compute the cost of plan we accumulate those values. Using this encoding significantly reduced the overhead since it is a straightforward implication between the disjunction of actions in a given step and indicates whether any of those actions is selected. Based on trails experiments, this encoding improved the solving time by at least a factor of 2. Another example is the goal predicate ordering ($\Upsilon$) encodings. Such a dimension can be encoded using bit-vectors such that each bit represents the ordering between two goal predicates, and the total order is captured by the whole value of the bit-vector. Although natural, this encoding caused a noticeable overhead. We used an alternative encoding via uninterpreted functions to capture the order between two goal appreciates and use the binary number encoding mentioned earlier to capture the total order. In comparison, this encoding had a lower overhead, with a performance improvement of around a factor of 3. We follow the same approach for the other dimensions to overcome any performance issues that any dimension may introduce.

\subsection{Forbid Behaviour \texorpdfstring{$\text{Iterative}_\texttt{X}$}{} Realisation\label{sec:fbireal}}

Since we are using SMT to implement $\texttt{FBI}_\texttt{X}$, we refer to this implementation as $\texttt{FBI}_\texttt{SMT}$. 
In this realisation, behaviour space is the conjunction of $\phi_n$ (\Cref{equ:planning-as-smt-equ}) and the encoded features $F_\Xi$.  

To realise $\operatorname{BehaviourGenerator}$ in \Cref{alg:fbi-planner-main} we make use of   $\operatorname{Encoder}(\Xi,n)=\phi^\prime_n$ (see \Cref{sec:behaviour-planning-realisation}) and 
four more functions: $\operatorname{IsSatisfiable}$, $\operatorname{ExtractModel}$, $\operatorname{PlanExtractor}$ and $\operatorname{EncodeFeatures}$. 
The first is used to determine whether a model $\mathcal{M}_{\phi_n^\prime}$ for  $\phi^\prime_n$ exists, the second and third to extract the model assignments, and a subset of those forming the plan $\pi$. The last one generates the encodings for the features defined.
The function 
$\operatorname{IsSatisfiable}(\phi_n^\prime)=t, t\in\{\top,\bot\}$ returns true if $\phi_n^\prime$ is satisfiable and false otherwise. If this returns true, it means that the  SMT solver has found a model for $\phi_n^\prime$, $\mathcal{M}_{\phi_n^\prime}$. We define $\operatorname{ExtractModel}(\phi^\prime_n)$ as a function that is used to extract $\mathcal{M}_{\phi_n^\prime}$. 
We use $\operatorname{PlanExtractor}(\mathcal{M}_{\phi_n^\prime})=\pi, \pi\in\Pi_\Xi$ to construct a plan from $\mathcal{M}_{\phi_n^\prime}$. $\operatorname{PlanExtractor}$ is a function that iterates over the model and extracts the enabled actions (i.e., the actions that are part of the plan) in every step to construct a plan. $\operatorname{EncodeFeatures}(F_\Xi)=\Theta$, where we use $\Theta$ to denote any conjunction of the provided features as described above (e.g., $\Theta\coloneq\Gamma\wedge\Upsilon$ or $\Theta\coloneq\Sigma_1\wedge\ldots\wedge\Sigma_n\wedge\Omega\wedge\Gamma$, etc). 
We refer to the updated formula as $\phi_n^\prime\coloneq\phi_n\wedge\Theta$. Note that any feature can be appended to $\phi_n$ as a logical formula.
%
%
%
The $\operatorname{BehaviourGenerator}$ can now be implemented using these functions.
%
%
To extract the behaviour from $\mathcal{M}_{\phi_n^\prime}$ implement $\operatorname{PBehaviour}$ (Def.\ \ref{def:plan-beh}) as  
$\operatorname{PBehaviour}(\odot_\Delta, \operatorname{PlanExtractor}(\mathcal{M}_{\phi_n^\prime}))=\langle \textit{Expr}_1=\odot_1(\pi),\ldots,\textit{Expr}_n=\odot_n(\pi)\rangle$
%
%
It is then easier for the user to understand what characteristics the plan has as its corresponding behaviour expresses them.


\Cref{alg:solver-behaviour} is an implementation for $\operatorname{BehaviourGenerator}$. First, it constructs $\phi_n^\prime$ of length $n$ using $\operatorname{Encoder}$. Then, it appends the behaviour space dimensions, constructed by $\operatorname{EncodeFeatures}$, to $\phi_n^\prime$. Finally, it forbids the plans’ behaviours in $\Psi_\Xi$ (Lines \ref{alg-line:construct-phi-n}-\ref{alg-line:forbid-behaviours}).  
Afterwards, it checks whether it can find a model for $\phi_n^\prime$. If it succeeds, then it extracts and returns the plan; otherwise, it returns an empty set (Lines \ref{alg-line:check-sat-start}-\ref{alg-line:check-sat-end}). 

\begin{algorithm}
    \caption{$\operatorname{BehaviourGenerator}$}\label{alg:solver-behaviour}
    \begin{algorithmic}[1]
        \REQUIRE $\Xi$: Planning task, $F_\Xi$: Diversity Features, $\Psi_\Xi$: Set of plans with different behaviours
        \ENSURE A plan ($\pi$) or an empty set ($\emptyset$) if no plan is found.
        \STATE $\text{;; compute $n$ which is the formula length.}$
        \STATE $\phi_n^\prime \gets \operatorname{Encoder}(\Xi, n) \wedge\operatorname{EncodeFeatures}(F_\Xi)$ \label{alg-line:construct-phi-n}
        \STATE $\odot_\Delta\gets \{\odot\ \vert\ \langle\Delta,\odot,\textit{Expr}\rangle \in F_\Xi\}$
        \STATE $\textit{behaviours}\gets\{\bigwedge_{e\in\operatorname{PBehaviour}(\odot_\Delta,\pi)} e\vert\pi\in\Psi_\Xi\}$ 
        \STATE $\phi^\prime_n\gets \phi^\prime_n  \bigwedge_{\mathcal{B}\in \textit{behaviours}} \neg\mathcal{B} $\label{alg-line:forbid-behaviours}
        \IF {$\operatorname{IsSatisfiable}(\phi_n^\prime)$} \label{alg-line:check-sat-start}
            \RETURN $\operatorname{PlanExtractor}( \operatorname{ExtractModel}(\phi_n^\prime))$
        \ENDIF
        \RETURN $\emptyset$ \label{alg-line:check-sat-end}
    \end{algorithmic}
\end{algorithm}

$\operatorname{PlanGenerator}$ used in \Cref{alg:fbi-k-planner-main} follows a similar implementation to \Cref{alg:solver-behaviour}. \Cref{alg:solver-plan} starts with encoding the planning problem into $\phi_n$ (Line \ref{alg-line:construct-phi-n-plan}). Afterwards, it forbids all plans in $\Psi_\Xi$, checks for satisfiability, and returns a plan $\pi$ if $\phi_n^\prime$ is satisfiable; otherwise, it is an empty set (Lines \ref{alg-line:plan-forbid-encode-start}-\ref{alg-line:forbid-check-sat-end}). A plan ($\pi$) is forbidden by setting its actions' boolean variables to false, then negating the Anding of this assignment. We use $\neg \pi$ to indicate this for simplicity.

\begin{algorithm}
    \caption{$\operatorname{PlanGenerator}$}\label{alg:solver-plan}
    \begin{algorithmic}[1]
        \REQUIRE $\Xi$: Planning task, $F_\Xi$: Diversity Features, $\Psi_\Xi$: Set of plans with different behaviours
        \ENSURE A plan ($\pi$) or an empty set ($\emptyset$) if no plan is found.
        \STATE $\text{;; compute $n$ which is the formula length.}$
        \STATE $\phi_n \gets \operatorname{Encoder}(\Xi, n)$ \label{alg-line:construct-phi-n-plan}
        \STATE $\phi_n \gets \phi_n \bigwedge_{\pi\in\Psi_\Xi} \neg\pi$ \label{alg-line:plan-forbid-encode-start} 
        \IF {$\operatorname{IsSatisfiable}(\phi_n)$} \label{alg-line:check-sat-start-plan}
            \RETURN $\operatorname{PlanExtractor}(\operatorname{ExtractModel}(\phi_n))$
        \ENDIF
        \RETURN $\emptyset$ \label{alg-line:forbid-check-sat-end}
    \end{algorithmic}
\end{algorithm}

\subsection{Illustrative Examples}

This section demonstrates some illustrative examples of our approach for the planning versions: classical, over-subscription and numeric. We used problem instance 1\footnote{https://github.com/AI-Planning/classical-domains/blob/main/classical/rovers/p01.pddl} for the classical version of the Rovers domain.  The optimal plan length to solve this instance is 10 actions. We used this length as the cost bound for the cost-bound dimension. Each planning problem is associated with its behaviour space $BS_\Delta$.

\paragraph{\textbf{Classical planning.}}
We utilised a behaviour space, denoted as $BS^{{\texttt{CLC}}}_\Delta$, which comprises two dimensions: (i) goal predicate ordering ($f_{go}$) and (ii) resource utilisation ($f_{ru}$) (i.e., $\Theta \coloneq \Upsilon\wedge\Lambda$). For the goal predicate ordering it does not require any additional information since it can be acquired from the problem instance. On the other hand, the resource utilisation requires additional information, which is the resource objects of interest (e.g., $\textit{AddInfo}_{ru}=\{\texttt{rover0},\texttt{rover1}\}$). Figures \ref{fig:behaviour-planning-rovers-plans-1}-\ref{fig:bspace-mapping-fi-vs-fbi} show the diverse plans generated for the classical planning problem version with behaviour space ($BS^{{\texttt{CLC}}}_\Delta$). \Cref{fig:behaviour-planning-rovers-plans-1} shows that $\texttt{FBI}_\texttt{SMT}$ can generate plans with different transmitted samples ordering using different numbers of rovers.

\begin{figure}
\subfigure[FBI diverse plans]{\label{fig:fbi-rovers-five-osp}\includegraphics[scale=0.6]{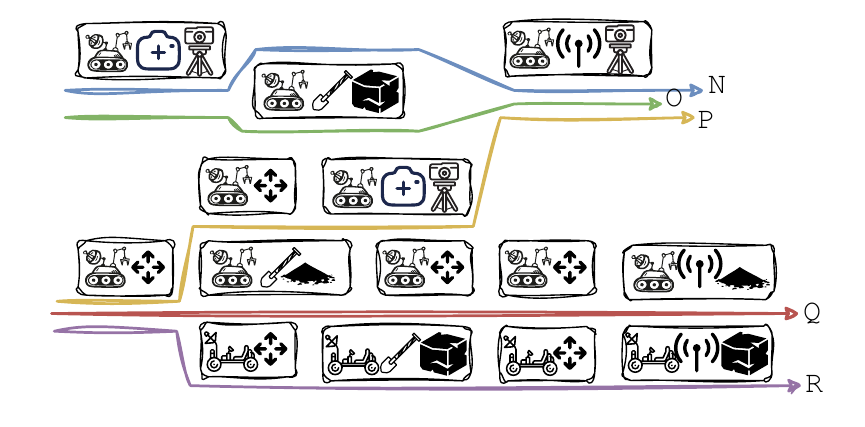}}
\subfigure[Behaviour space]{\label{fig:fbi-rovers-five-osp-bspace}\includegraphics[scale=0.7]{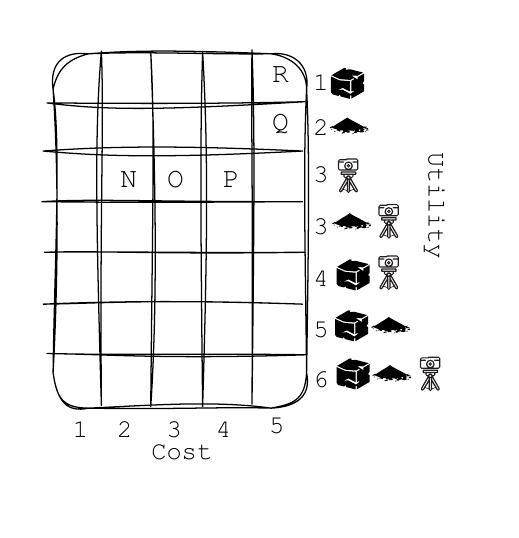}}
\caption{Diverse plans generated for the rover planning problem OSP version. Letters P, Q, R, O and N are the plans in \Cref{fig:fbi-rovers-five-osp}. Some actions are removed in favour of illustration. See supplementary materials for full plans.}\label{fig:fbi-rovers-plans-bspace-osp}%
\end{figure}

\paragraph{\textbf{Over-subscription planning.}} 
We utilised a behaviour space, denoted as $BS_\Delta^{{\texttt{OSP}}}$, which comprises two dimensions: (i) cost bound ($f_{cb}$) and (ii) utility value ($f_{uv}$) (i.e., $\Theta \coloneq \Gamma\wedge\Omega$). For the cost bound, we set the limit to half the optimal plan length (i.e., 5 actions). Regarding the utility value dimension ($f_{uv}$), we assigned the following utilities to the transmitted samples: soil has a utility of 1, rock has a utility of 2, and communicating an image has a utility of 3 (i.e., $\{\langle\texttt{S},1\rangle, \langle\texttt{R},2\rangle, \langle\texttt{I},3\rangle\}$). %
\Cref{fig:fbi-rovers-plans-bspace-osp} shows five plans for the rover domain planning problem, with the cost bound set to half of the optimal length. \Cref{fig:fbi-rovers-five-osp-bspace} maps these plans into the behaviour space. \Cref{fig:fbi-rovers-plans-bspace-osp} illustrates that behaviour planning can produce plans with varying costs and utility values. For instance, plans \texttt{R} and \texttt{Q} have the same cost but differ in their utilities; one plan transmits both soil and rock samples, while the other transmits only the soil data. The used behaviour space can explain plans in terms of cost and utility, similar to \citet{explain-osp}'s work. Furthermore, it captures the essence of net-benefit planning, maximising the difference between cost and utility~(\cite{net-benefit}). Behaviour planning could capture such information by extending the behaviour space with another dimension: the difference between the utility and cost dimensions. 

\begin{figure}
\centering   
\subfigure[FBI diverse plans]{\label{fig:fbi-rovers-five-numeric}\includegraphics[width=\linewidth]{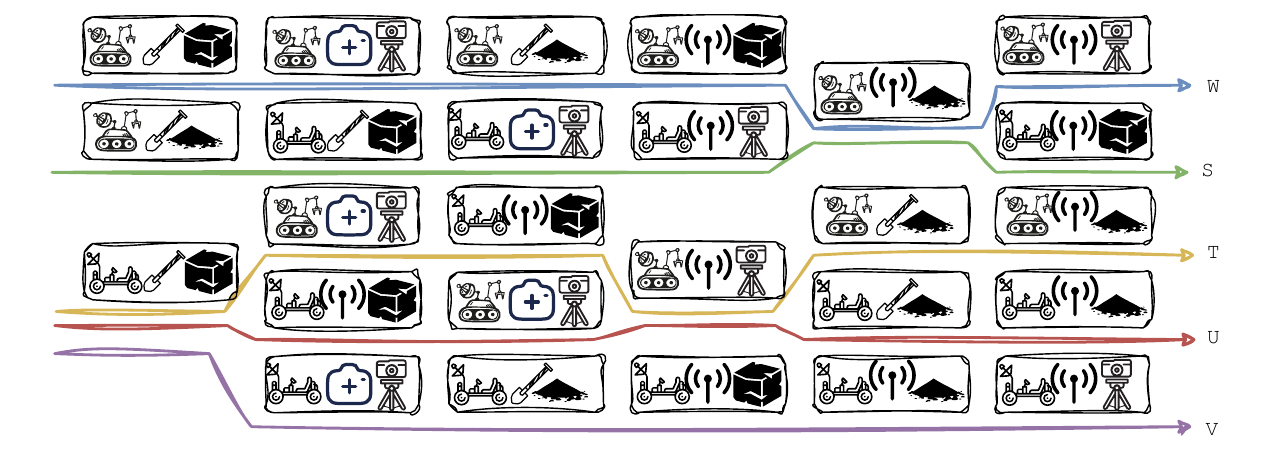}}

\subfigure[Behaviour space]{\label{fig:fbi-rovers-five-numeric-bspace}\includegraphics[scale=0.55]{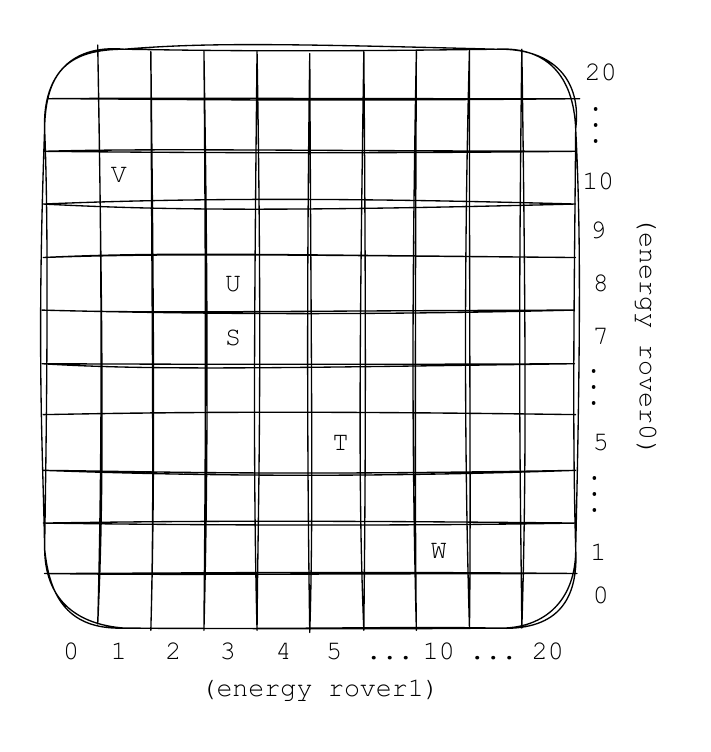}}
\caption{Diverse plans generated for the numeric version of the Rovers planning problem. Letters S, T, U, V and W are the plans in \Cref{fig:fbi-rovers-five-numeric}. Some actions are removed in favour of illustration. See supplementary materials for full plans.}\label{fig:fbi-rovers-plans-bspace-numeric}%
\end{figure}

\paragraph{\textbf{Numeric planning.}} 
For the planning problem version\footnote{https://github.com/pyPMT/numeric-domains/blob/main/rover/domain.pddl}, we utilised a behaviour space, denoted as $BS_\Delta^{\texttt{numeric}}$, with two numeric fluent features ($f_{\mathit{nf}}^\texttt{rover0}, f_{\mathit{nf}}^\texttt{rover1}$) and the cost bound  set to optimal length (i.e., $\Theta\coloneq\Sigma_{\texttt{rover0}}\wedge\Sigma_{\texttt{rover1}}$). 
Those two numeric fluents dimensions contains two numeric fluents that reflect the rovers' energy. The relevant information is given by $\textit{AddInfo}_{\mathit{nf}}^\texttt{rover0}=\langle \texttt{energy\_rover0}, 0, 100, 5 \rangle$, and $\textit{AddInfo}_{\mathit{nf}}^\texttt{rover1}=\langle \texttt{energy\_rover1}, 0, 100, 5 \rangle$.
%
\Cref{fig:fbi-rovers-five-numeric} illustrates five plans created for the planning problem. Conversely, \Cref{fig:fbi-rovers-five-numeric-bspace} maps these plans into the behaviour space. As we can infer from the behaviour space, all plans use the rovers in different sample gathering tasks, which results in different energy levels. 
$box_{\texttt{energy\_rover0}}=0$ means that a \texttt{rover0}'s energy range is between 0 and 5, and it keeps incrementing based on the step size provided in $\textit{AddInfo}_{\mathit{nf}}^\texttt{rover0}, \textit{AddInfo}_{\mathit{nf}}^\texttt{rover1}$ for each rover. 

Thanks to behaviour space, users can use it to generate diverse plans based on their customised diversity model. Finally, behaviour planning is the first diverse approach that supports other planning variants: over-subscription and numeric planning. 

\section{Experimental Setup \& Evaluation}\label{sec:exp-diss}

In this section, we present a series of experiments designed to demonstrate the validity of behaviour planning for diverse planning and discuss its advantages compared to other diverse planning frameworks.
Here, we describe the experimental setup, research question, and experiment results. The findings subsection covers limitations of our framework implementation performance and suggestions for addressing them. 
In our experiments, we answer the research question (Q -- How effectively can behaviour planning produce diverse plans for different planning paradigms?). 

\subsection{Experiment Setup}
We implemented behaviour planning using Python and the Z3 SMT solver~(\cite{z3-solver}), which is made publicly available\footnote{\url{https://github.com/MFaisalZaki/pyBehaviourPlanningSMT.git}}.
To answer our research question, we examine the capabilities of our behaviour planning implementation to generate diverse plans for different planning categories.
In this experiment, we assume makespan optimal planning. 

Our experiments are configured as follows: we solved a set of planning tasks on an AMD EPYC 7763 64-Core Processor@2.4GHz for $k$ plans, where $k\in\{5,10,100,1000\}$ and relative quality bounds (i.e. $c{=}\operatorname{round}(q*l)$, where $l$ is the optimal length), where $q=1.0$ implies the generated set $\Psi_\Xi$ has optimal solutions only. On the other hand, $q=2.0$ indicates that $\Psi_\Xi$ contains plans with cost more than the optimal value but less than double the optimal value. Each task is solved once for every configuration (i.e., $q$, $k$, planner). After generating $k$ plans, we compute the behaviour count for those $k$ plans. We restrict the resources used by each task to one CPU core, 30 minutes and 16 GB of memory. 
Regarding the resource utilisation utility value and numeric fluent information, we provided $\textit{AddInfo}_{ru}$ and $\textit{AddInfo}_\mathit{nf}$ through external files since PDDL does not allow us to provide such information.

We discuss our results in relation to those obtained with \texttt{FI} and \texttt{Symk}. 
In our experiments, \texttt{FI} assumes that it is used in conjunction with \citeauthor{vadlamudi2016combinatorial}'s framework. Such an assumption makes \texttt{FI} ignore the $k$ constraint, generating substantially more plans than requested. To have comparable results between \texttt{FI} and $\texttt{FBI}_\texttt{SMT}$, we have used three configurations for \texttt{FI}. The $\texttt{FI}_{\texttt{max}}$ configuration uses \texttt{MaxSum} model with the stability metric\footnote{\citet{srivastava2007domain} defined the stability metric as the Jaccard measurement between two plans' actions. } to select $k$ plans. The $\texttt{FI}_\texttt{k}$ configuration selects the first $k$ unique plans generated by \texttt{FI}. Finally, $\texttt{FI}_\texttt{bc}$ selects $k$ plans that maximise behaviour count. \texttt{FI} and \texttt{Symk} are based on Fast Downward~(\cite{helmert2006fast}), which limits them to solving classical planning problems. A version of \texttt{Symk} supports over-subscription planning problems and, therefore, we can compare it with $\texttt{FBI}_\texttt{SMT}$. No diverse or top-k planners are currently available for numeric planning. Hence, we compare $\texttt{FBI}_\texttt{SMT}$ with its naive version that forbids plans only.

To answer our research question (Q -- How effectively can behaviour planning produce diverse plans for different types of planning?), we test behaviour planning with three types of planning variants: (i) classical, (ii) over-subscription and (iii) numerical, each planning variant has its unique behaviour space. We computed the behaviour count and coverage for all planners (i.e., $\texttt{FI}_\texttt{bc/k/max}$ and $\texttt{FBI}_\texttt{SMT}$) that solved those tasks in each planning class. To evaluate differences in behaviour count, we computed the statistical significance between each planner (i.e., $\texttt{FI}_\texttt{k/bc/max}$) and $\texttt{FBI}_\texttt{SMT}$ using a vector of behaviour count pairs per commonly solved instance per $k$ and $q$ values.

\subsection{Evaluation}
Here we report the results of our experiments for each planning variant: classical, over-subscription and numerical.  

\begingroup
\setlength{\tabcolsep}{3.5pt}
\renewcommand{\arraystretch}{1.1} 
\begin{table}
\caption{The behaviour space column shows which dimensions are used to compute the behaviour count. The coverage columns show the number of solved instances by $\texttt{FI}_\texttt{k/bc/max}$ and $\texttt{FBI}_\texttt{SMT}$ for various values of $k$ with different relative quality bounds $q$. For coverage, bold highlights the higher number of solved instances.  CI denotes the number of commonly solved instances by the planners. The behaviour count columns denote the accumulated behaviour counts for different values of quality bounds $q$ and the number of extracted plans $k$ for each planner computed based on the common instances solved by planners.  For behaviour count, bold highlights statistical significance differences for \texttt{FI} vs $\texttt{FBI}_\texttt{SMT}$.}\label{tbl:ttest-fbi-fi}
\centering
\begin{tabular}{ccl|cccc|c|cccc}
\hline
\multirow{2}{*}{\begin{tabular}[c]{@{}c@{}}$F_\Xi$\end{tabular}} & \multirow{2}{*}{q}   & \multirow{2}{*}{k} & \multicolumn{4}{c|}{Coverage}                                                    & \multirow{2}{*}{CI} & \multicolumn{4}{c}{Behaviour Count}                                              \\
                                                                           &                      &                    & $\texttt{FI}_{\texttt{bc}}$ & $\texttt{FI}_{\texttt{max}}$ & $\texttt{FI}_{\texttt{k}}$ & $\texttt{FBI}_\texttt{SMT}$ &                     & $\texttt{FI}_{\texttt{bc}}$ & $\texttt{FI}_{\texttt{max}}$ & $\texttt{FI}_{\texttt{k}}$ & $\texttt{FBI}_\texttt{SMT}$ \\ \hline
\multirow{4}{*}{$\{f_{go},f_{ru}\}$}                                 & \multirow{4}{*}{1.0} & 5                  & 535                & \textbf{544}           & \textbf{544}      & 251            & 212                 & 549                & 520                    & 516               & \textbf{829}   \\
                                                                           &                      & 10                 & 445                & \textbf{454}           & \textbf{454}      & 221            & 164                 & 756                & 700                    & 696               & \textbf{1272}  \\
                                                                           &                      & 100                & 249                & \textbf{257}           & \textbf{257}      & 178            & 87                  & 2772               & 2775                   & 2670              & \textbf{6287}  \\
                                                                           &                      & 1000               & 86                 & 88            & 88       & \textbf{142}            & 45                  & 8161               & 8158                   & 8152              & \textbf{41055} \\ \hline \hline
\multirow{4}{*}{$\{f_{go},f_{ru},f_{cb}\}$}                   & \multirow{4}{*}{2.0} & 5                  & 580                & \textbf{589}           & \textbf{589}      & 196            & 187                 & 566                & 534                    & 533               & \textbf{874}   \\
                                                                           &                      & 10                 & 542                & \textbf{551}           & \textbf{551}      & 195            & 183                 & 940                & 879                    & 884               & \textbf{1680}  \\
                                                                           &                      & 100                & 375                & \textbf{383}           & \textbf{383}      & 193            & 155                 & 3496               & 3402                   & 3352              & \textbf{11765} \\
                                                                           &                      & 1000               & 114                & 114                    & 115      & \textbf{192}            & 71                  & 10917              & 10675                  & 10779             & \textbf{48491} \\ \hline
\end{tabular}
\end{table}
\endgroup

\begin{figure}
\subfigure[$q=1.0$]{\includegraphics[scale=0.31]{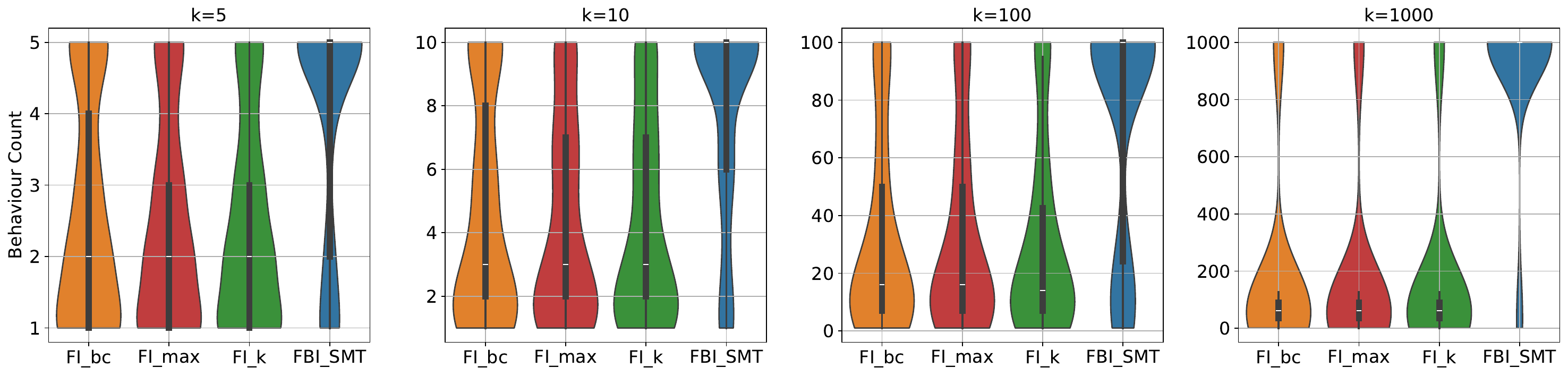}}
\subfigure[$q=2.0$]{\includegraphics[scale=0.31]{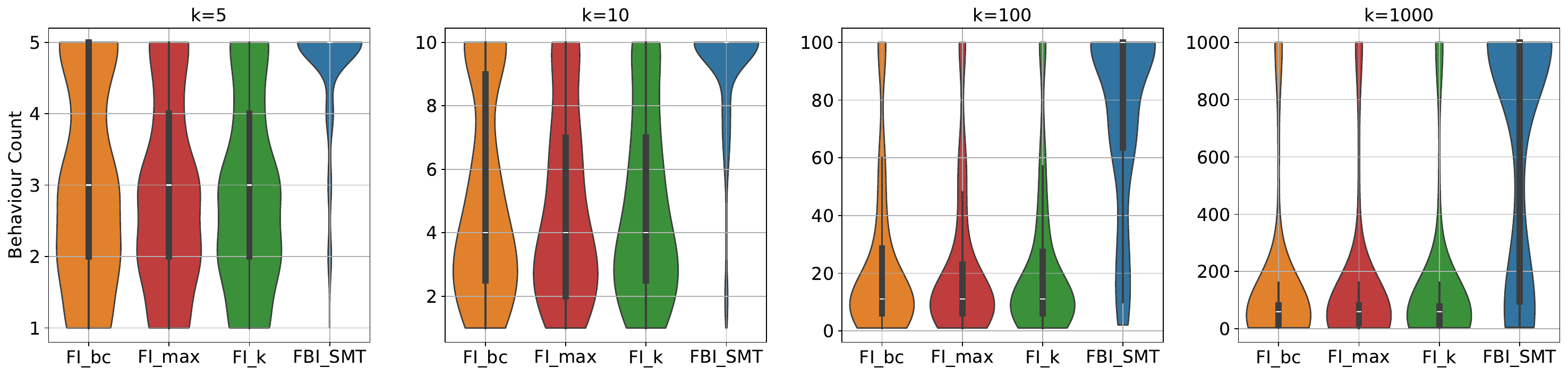}}
\caption{Violin plots for classical planning problems experiments. Planners are ordered as follows: $\texttt{FI}_\texttt{bc}, \texttt{FI}_\texttt{max}, \texttt{FI}_\texttt{k}, \texttt{FBI}_\texttt{SMT}$.}\label{fig:violin-classical-results}
\end{figure}

\paragraph{\textbf{Classical planning.}} In this setup, we solved 53 domains, 1278 tasks\footnote{\url{https://github.com/AI-Planning/classical-domains}}  using $\texttt{FI}_\texttt{k/bc/max}$ and $\texttt{FBI}_\texttt{SMT}$ for different $k\in\{5,10,100,1000\}$ and $q\in\{1.0,2.0\}$. We used a 3-dimensional behaviour space including cost bound $f_{cb}$, resource utilisation $f_{ru}$, and goal predicate ordering $f_{go}$, where $\Theta:= \Gamma \wedge \Lambda \wedge \Upsilon$, as described in \Cref{sec:behaviour-planning-realisation}.   For quality bound 1.0, there is no point in adding the cost-bound dimension since all generated plans will have the exact cost.
Then, we compared the behaviour counts of each planner based on the commonly solved instances. \Cref{tbl:ttest-fbi-fi} shows the accumulated coverage and behaviour count for the commonly solved instances by the planners. From this table, we can conclude that $\texttt{FBI}_\texttt{SMT}$ generates more diverse plans compared to \texttt{FI}. 
\Cref{tbl:ttest-fbi-fi} demonstrates that $\texttt{FBI}_\texttt{SMT}$ outperforms all \texttt{FI}’s configurations in generating more diverse plans with statistical significance ($p<0.05$) for all values of $k$. \Cref{fig:violin-classical-results} presents the distribution of the behaviour counts per $q$ and $k$ from \Cref{tbl:ttest-fbi-fi}. From \Cref{tbl:ttest-fbi-fi} it is obvious that $\texttt{FBI}_\texttt{SMT}$ generated more diverse plans compared to $\texttt{FI}_\texttt{bc/k/max}$. This was made possible because $\texttt{FBI}_\texttt{SMT}$ incorporates the diversity model during the planning process unlike the current approach which does not account for any diversity model during planning.

\begin{figure}
\centering
\subfigure[$q=0.25$]{\includegraphics[scale=0.3]{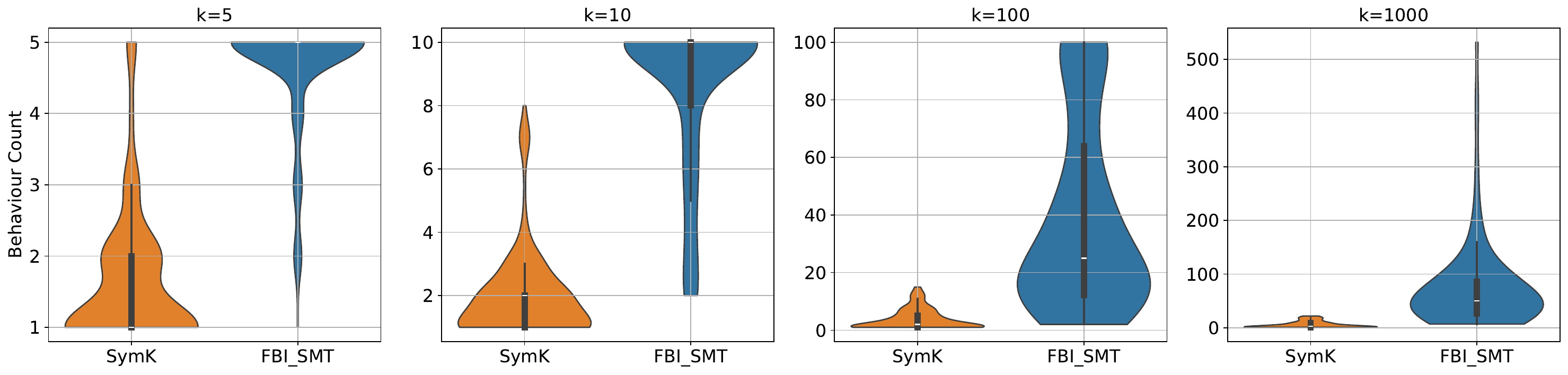}}
\subfigure[$q=0.5$]{\includegraphics[scale=0.3]{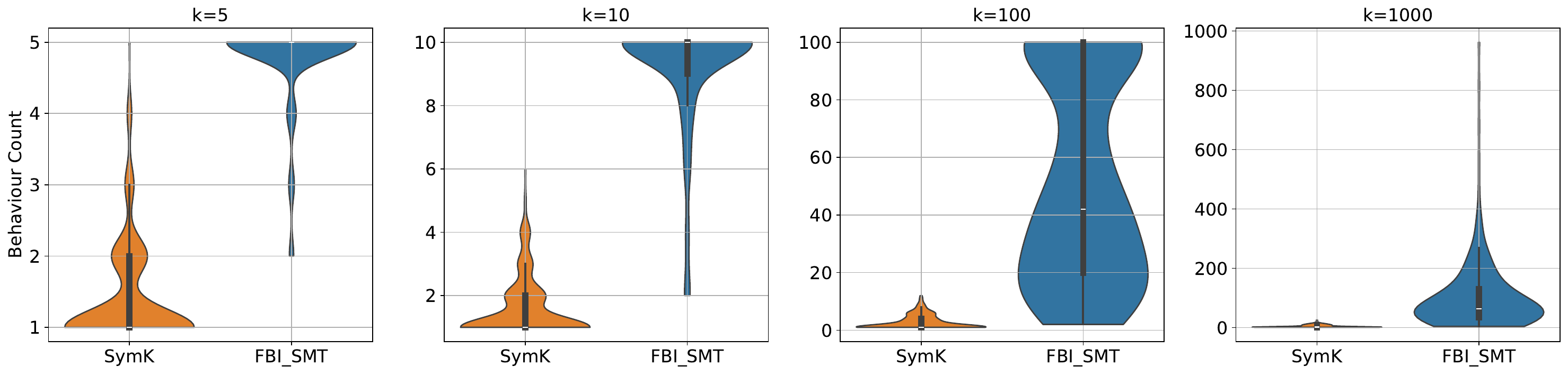}}
\subfigure[$q=0.75$]{\includegraphics[scale=0.3]{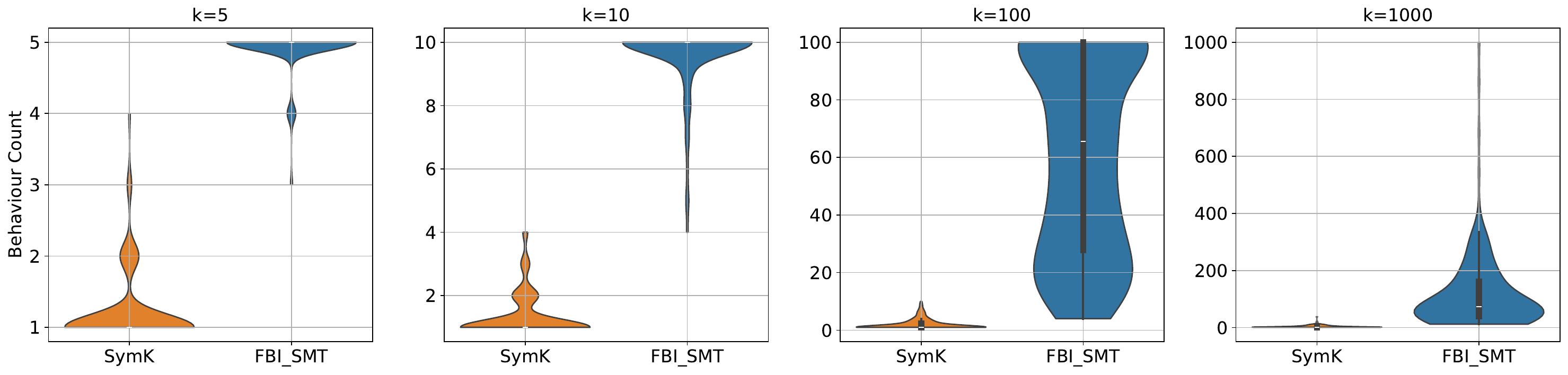}}
\subfigure[$q=1.0$]{\includegraphics[scale=0.3]{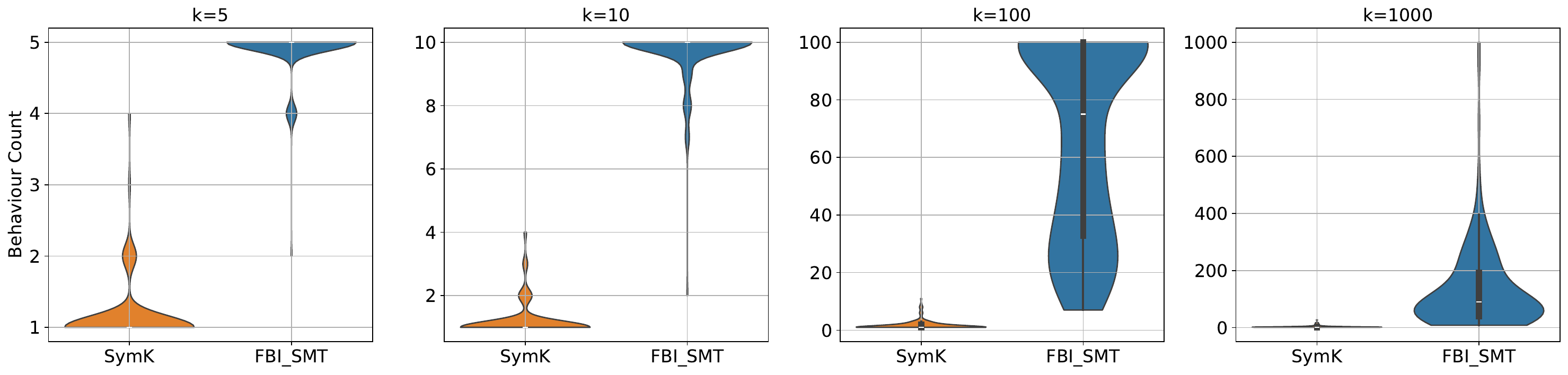}}
\caption{Violin plots for OSP planning problems experiments. Planners are ordered as follows: $\texttt{SymK}, \texttt{FBI}_\texttt{SMT}$.} \label{fig:violin-osp-results}

\end{figure}

\paragraph{\textbf{Over-subscription planning.}} In this setup, we transformed the classical planning tasks used in the previous experiment into over-subscription planning ones by incrementally assigning utility values for goal predicates and converting the goal state from a conjunction of predicates to a disjunction. We solve those updated tasks using \texttt{SymK} and $\texttt{FBI}_\texttt{SMT}$ for different quality bounds (i.e., $q=\{0.25,0.5,0.75,1.0\}$) following the same evaluation used by \citet{speck-katz-aaai2021}. Regarding the behaviour space, we used two-dimensional space: one for cost $f_{cb}$ while the second for utility $f_{uv}$, where $\Theta:=\Omega\wedge\Lambda$. 
As mentioned above, there are no planners except for \texttt{SymK} that can generate a set of plans for over-subscription planning tasks~(\cite{speck-katz-aaai2021}), hence we used it as a baseline in this planning category. \Cref{table:osp-results} shows the behaviour count, mean, and standard deviation for the commonly solved instances by the planners. Like classical planning, $\texttt{FBI}_\texttt{SMT}$ shows that it can generate diverse plans for the over-subscription task with statistical significance ($p<0.05$) for all values of $k$ compared to \texttt{Symk}. \Cref{fig:violin-osp-results} presents the distribution of behaviour counts from \Cref{table:osp-results}.

\begin{table}
\caption{Coverage and behaviour count results for the over-subscription planning tasks, same structure used in previous tables.}\label{table:osp-results}
\centering
\begin{tabular}{ccl|cc|c|cc}
\hline
\multirow{2}{*}{\begin{tabular}[c]{@{}c@{}}$F_\Xi$\end{tabular}} & \multirow{2}{*}{q}    & \multirow{2}{*}{k} & \multicolumn{2}{c|}{Coverage}    & \multirow{2}{*}{CI} & \multicolumn{2}{c}{Behaviour Count} \\
                                                                           &                       &                    & $\texttt{Symk}$ & $\texttt{FBI}_\texttt{SMT}$ &                     & $\texttt{Symk}$   & $\texttt{FBI}_\texttt{SMT}$  \\ \hline
\multirow{16}{*}{$\{f_{uv},f_{cb}\}$}                                   & \multirow{4}{*}{0.25} & 5                  & \textbf{522}    & 194            & 187                 & 304               & \textbf{871}    \\
                                                                           &                       & 10                 & \textbf{490}    & 185            & 176                 & 346               & \textbf{1517}   \\
                                                                           &                       & 100                & \textbf{422}    & 164            & 154                 & 579               & \textbf{6059}   \\
                                                                           &                       & 1000               & \textbf{197}    & 135            & 93                  & 548               & \textbf{7250}   \\ \cline{2-8} \cline{2-8} \cline{2-8} \cline{2-8} 
                                                                           & \multirow{4}{*}{0.5}  & 5                  & \textbf{560}    & 247            & 240                 & 334               & \textbf{1146}   \\
                                                                           &                       & 10                 & \textbf{555}    & 243            & 235                 & 368               & \textbf{2155}   \\
                                                                           &                       & 100                & \textbf{523}    & 233            & 221                 & 609               & \textbf{11624}  \\
                                                                           &                       & 1000               & \textbf{339}    & 217            & 173                 & 752               & \textbf{18705}  \\ \cline{2-8} \cline{2-8} \cline{2-8} \cline{2-8} 
                                                                           & \multirow{4}{*}{0.75} & 5                  & \textbf{519}    & 227            & 213                 & 257               & \textbf{1046}   \\
                                                                           &                       & 10                 & \textbf{519}    & 226            & 212                 & 285               & \textbf{2035}   \\
                                                                           &                       & 100                & \textbf{501}    & 224            & 208                 & 429               & \textbf{12674}  \\
                                                                           &                       & 1000               & \textbf{373}    & 219            & 185                 & 654               & \textbf{21891}  \\ \cline{2-8} \cline{2-8} \cline{2-8} \cline{2-8}  
                                                                           & \multirow{4}{*}{1.0}  & 5                  & \textbf{493}    & 210            & 190                 & 218               & \textbf{933}    \\
                                                                           &                       & 10                 & \textbf{491}    & 210            & 190                 & 228               & \textbf{1846}   \\
                                                                           &                       & 100                & \textbf{484}    & 209            & 188                 & 320               & \textbf{12663}  \\
                                                                           &                       & 1000               & \textbf{364}    & 202            & 174                 & 454               & \textbf{23567}  \\ \hline
\end{tabular}
\end{table}

\begin{figure}
\subfigure[$q=1.0$]{\includegraphics[scale=0.3]{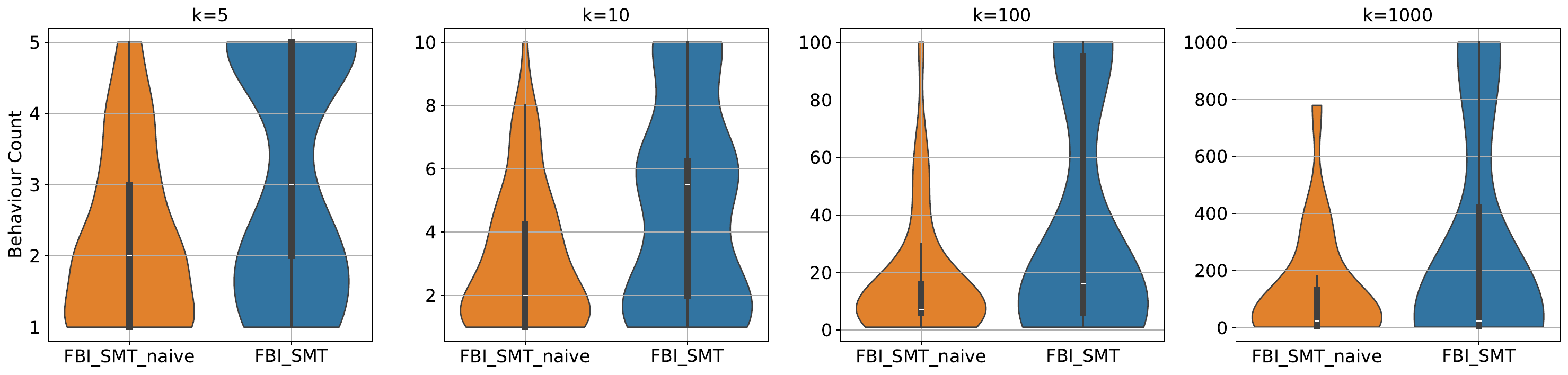}}
\subfigure[$q=2.0$]{\includegraphics[scale=0.3]{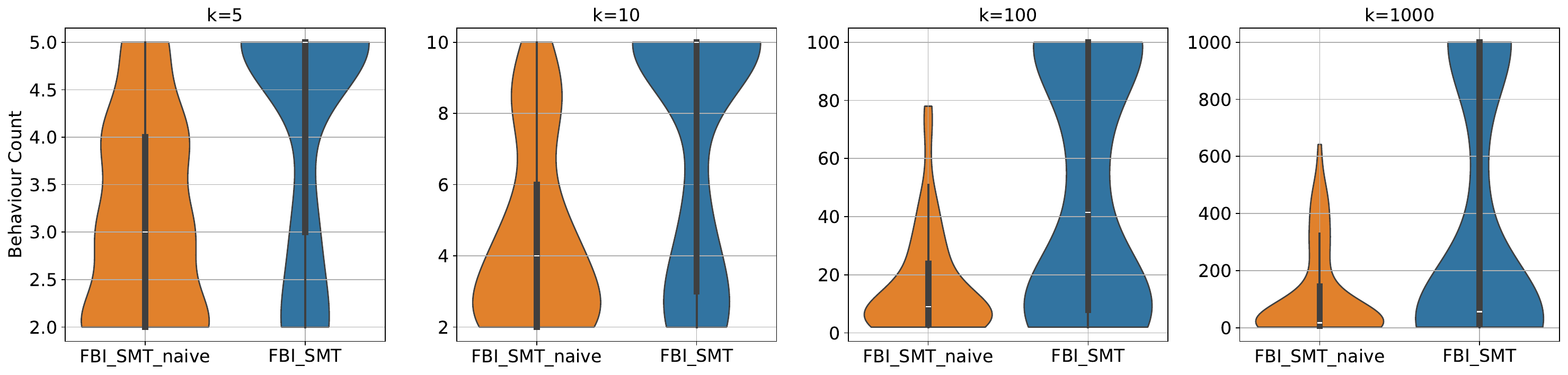}}
\caption{Violin plots for numeric planning problems experiments. Planners are ordered as follows: $ \texttt{FBI}^\texttt{NAIVE}_\texttt{SMT}, \texttt{FBI}_\texttt{SMT}$.}\label{fig:violin-numeric-results}
\end{figure}

\paragraph{\textbf{Numerical planning.}} In this setup, we solved numerical planning tasks (28 domains, 982 tasks)\footnote{\url{https://github.com/pyPMT/numeric-domains.git}} for different $k\in\{5,10,100,1000\}$ and $q\in\{1.0,2.0\}$ values. We used a three-dimensional behaviour space: cost-bound $f_{cb}$, goal predicate ordering $f_{go}$ and numeric fluent $f_{nf}$, where  $\Theta:= \Upsilon \wedge\Sigma_1\wedge\ldots\wedge\Sigma_n\wedge \Gamma$. The dimensions of numeric fluents depend on the problem instance. For example, the rover's third instance has two fluents named $\texttt{energy\_rover0}$ and $\texttt{energy\_rover1}$. We created $\Sigma_\texttt{energy\_rover0}$ and $\Sigma_\texttt{energy\_rover1}$ to represent each fluent.  
Unfortunately, no planners are available to generate a set of plans for this planning type. This makes $\texttt{FBI}_\texttt{SMT}$ the first planner that supports diverse numerical planning. To have a baseline, we compare $\texttt{FBI}_\texttt{SMT}$ against  $\texttt{FBI}_\texttt{SMT}^\texttt{naive}$, a $\texttt{FBI}_\texttt{SMT}$ that generates plans based on a behaviour space with no dimensions (i.e., $\Theta\coloneq\top$). \Cref{table:numeric-results} shows that $\texttt{FBI}_\texttt{SMT}$ can generate diverse plans for numerical planning tasks with statistical significance ($p<0.05$) for all values of $k$ compared to its naive version. \Cref{fig:violin-numeric-results} presents the distribution of the behaviour counts  from \Cref{table:numeric-results}.

\begin{table}
\caption{Coverage and behaviour count results for the numerical planning tasks, we follow the structure used in previous tables.}\label{table:numeric-results}
\centering
\begin{tabular}{c|c|l|cc|c|cc}
\hline
\multirow{2}{*}{$F_\Xi$}                                  & \multirow{2}{*}{q}   & \multirow{2}{*}{k} & 
\multicolumn{2}{c|}{Coverage}                                             & \multirow{2}{*}{CI} & \multicolumn{2}{c}{Behaviour Count}                                      \\
                                                        &                      &                    & $\texttt{FBI}_\texttt{SMT}$ & $\texttt{FBI}_\texttt{SMT}^\texttt{naive}$ &                     & $\texttt{FBI}_\texttt{SMT}$ & $\texttt{FBI}_\texttt{SMT}^\texttt{naive}$ \\ \hline
\multirow{4}{*}{$\{f_{go},f_{nf}\}$}               & \multirow{4}{*}{1.0} & 5                  & $\textbf{88}$                          & $\textbf{88}$                                         & 85                  & $\textbf{269}$                         & 189                                        \\
                                                        &                      & 10                 & $\textbf{79}$                          & $\textbf{79}$                                         & 76                  & $\textbf{367}$                         & 238                                        \\
                                                        &                      & 100                & 36                          & $\textbf{38}$                                         & 35                  & $\textbf{1289}$                        & 599                                        \\
                                                        &                      & 1000               & 20                          & $\textbf{22}$                                         & 19                  & $\textbf{5271}$                        & 2465                                       \\ \hline \hline
\multirow{4}{*}{$\{f_{go},f_{nf},f_{cb}\}$} & \multirow{4}{*}{2.0} & 5                  & 87                          & $\textbf{96}$                                         & 87                  & $\textbf{355}$                         & 272                                        \\
                                                        &                      & 10                 & 87                          & $\textbf{96}$                                         & 87                  & $\textbf{628}$                         & 397                                        \\
                                                        &                      & 100                & 81                          & $\textbf{88}$                                         & 78                  & $\textbf{3901}$                        & 1278                                       \\
                                                        &                      & 1000               & 75                          & $\textbf{84}$                                         & 75                  & $\textbf{26776}$                       & 8147                                       \\ \hline \hline
\end{tabular}
\end{table}

\section{Discussion}\label{sec:discussion}
In this section, we discuss the results of our experiments conducted in \Cref{sec:exp-diss}. Subsequently, we demonstrate how various users can leverage behaviour planning and explore the advantages it offers. Additionally, we delve into other scenarios where behaviour planning can be effectively employed.

\begin{figure}[!thp]
  \centering
  \subfigure[\texttt{FI} plans.]{\label{fig:fi-rovers-5-plans}
    \includegraphics[width=0.9\linewidth]{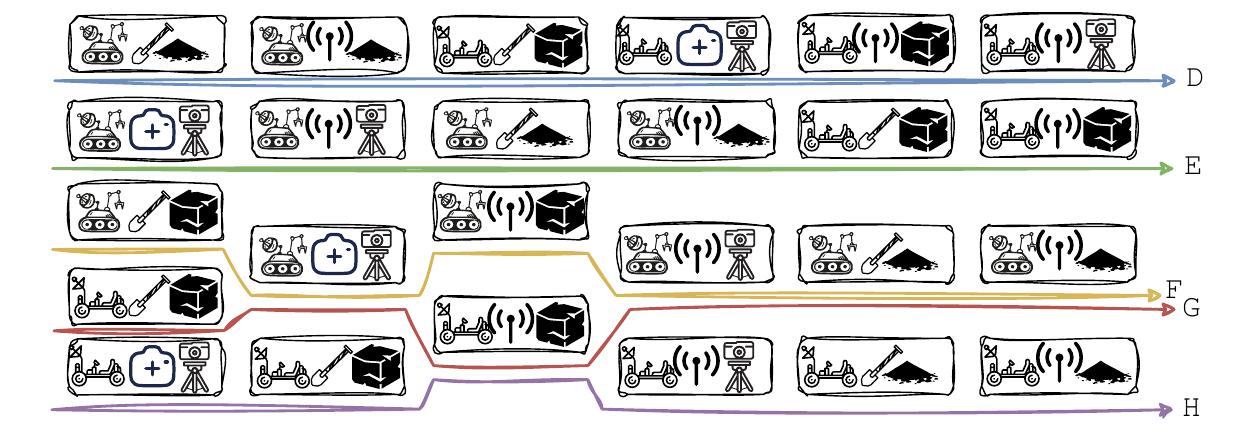}}
  \subfigure[$\texttt{FBI}_\texttt{SMT}$ plans.]{\label{fig:behaviour-planning-rovers-plans-1}\includegraphics[width=0.9\linewidth]{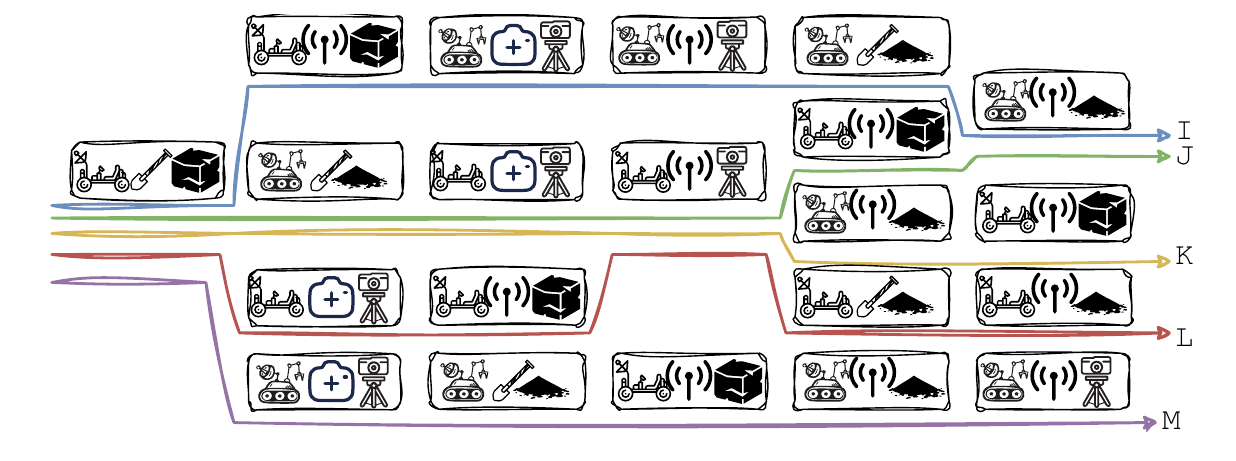}}
  \subfigure[Behaviour space]{\label{fig:bspace-mapping-fi-vs-fbi}\includegraphics[scale=0.75]{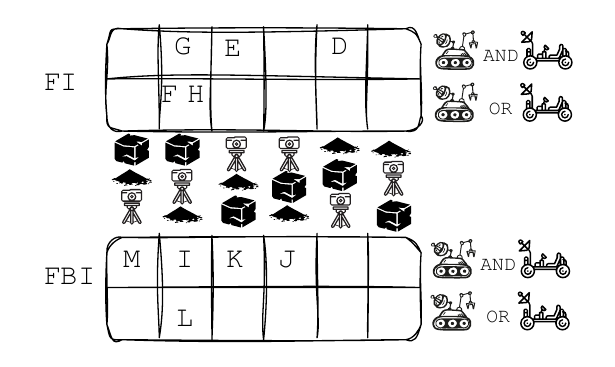}}
  \caption{Illustrations of plans generated by \texttt{FI} and $\texttt{FBI}_\texttt{SMT}$ for the Mars rover planning task. Some actions are removed in favour of illustration. See supplementary materials for full plans.}
  \label{fig:rovers-subfigures}
\end{figure}

\textbf{Findings.} The expressivity offered by planning-as-satisfiability comes with costs. One of those costs is lower coverage compared to other planners. The experiments show that \texttt{Symk} has the highest coverage, thanks to its symbolic search approach, but the diverse planner \texttt{FI} is also able to find more plans compared to $\texttt{FBI}_\texttt{SMT}$, particularly at lower values of $k$. 
We believe one of the reasons behind this lower coverage is the resources consumed to ground the actions. Grounding the actions is required by the used encodings, and it is a costly operation for a subset of the domains. Thus, using a lifted encoding would give the solver more time to search for plans, which, in turn, is expected to improve coverage. A well-known technique to improve the performance of planning-as-satisfiability approaches is through parallelism (i.e., selection of multiple actions at the same step) to reduce the impact of the plan length on the planner's performance. However, this would impact the optimality of the plan since this kind of encoding tends to enable several non-conflicting actions simultaneously to find a plan quickly. Yet, this can be avoided by extending the encoding to account for optimality~(\cite{Leofante23}).
Besides the grounding factor, the plan length directly impacts the coverage. When the plan length increases, the number of possible combinations when selecting actions per step also increases, making it more time-consuming. On the bright side, this increases the possibility of having more behaviours. 
In summary, our behaviour planning framework demonstrated its ability to generate more diverse plans compared to the currently employed framework. Moreover, behaviour planning can produce diverse plans based on a specified diversity model and supports various planning categories, unlike the current framework, which fails to consider the diversity model. To demonstrate the advantages of adopting a diversity model during planning and restricting plans’ properties (i.e., behaviours) rather than their reordering, Figures \ref{fig:fi-rovers-5-plans} and \ref{fig:behaviour-planning-rovers-plans-1} showcase five diverse plans generated by \texttt{FI} and $\texttt{FBI}_\texttt{SMT}$ for the motivating example discussed earlier. We observe that when we map the \texttt{FI} generated plans into the problem’s behaviour space (\Cref{fig:bspace-mapping-fi-vs-fbi}), we find that the plans \texttt{F} and \texttt{H} are distinct. Conversely, both plans utilise the same number of resources and transmit the gathered samples in the same order.

\paragraph{\textbf{Usage.}}
Choosing which diversity planning formulation to use depends on the end user’s requirements. If the user perceives plans as distinct on a structural level, it would be more appropriate to use the older problem formulation. This is because two symmetrical plans are considered identical based on the behaviour space, which contradicts the user’s needs. Conversely, if the user requires to distinguish between plans on a semantic level, they can use our new formulation and utilise the behaviour space to describe their diversity model. 
Recall the Mars planning problem, where astronauts distinguish between plans based on two features: the order of transmitted samples and the rovers used. These features were modelled using a two-dimensional behaviour space, where each dimension represented a feature. This space was then used to generate plans based on the information provided. This diversity model was possible because the suggested reformulation of the planning problem allowed encoding the factors into an n-dimensional grid. 
In case the user still uses our new formulation and wants to differentiate between plans on a structural level, they need to make sure that each behaviour contains only one plan. One way to achieve this is to use a dimension that considers the conjunction of a plan's actions as a unique value for this dimension. However, doing this would increase the behaviour space size and would introduce more challenges to the planner to find plans.

\paragraph{\textbf{Benefits.}} 
Users in the motivating example are three distinct personas: an astronaut (end user), a scientist (domain expert), and a planning expert (algorithm designer). Each user will benefit from behaviour planning differently. For instance, in the Mars rover context, the astronaut will provide their description of diversity.
These plans will have distinct characteristics, such as sample order and resource usage. Based on their specific situation, the astronaut will select plans that meet their needs.
The scientist would benefit from behaviour planning regarding modelling complex diversity representation. Using older frameworks, they would need to encode diversity into a single numeric function, which can be challenging to encode some features into a numeric function. However, using \texttt{BSS}, they can simply describe the diversity model through an n-dimensional grid, where each dimension denotes a feature of interest.
The planning expert will either implement new dimensions or reuse the ones from \texttt{BSS} SMT realisation based on the scientist’s requirements. Then, they can use $\texttt{FBI}_\texttt{SMT}$ to generate diverse plans based on a provided behaviour space.

\paragraph{\textbf{Variability.}} Behaviour planning can be a proxy for preference-based planning~(\cite{jorge2008planning}), where not all preferences are known beforehand. For example, the astronaut in the motivating example may want to see all possible plans with different transmission order and rover usage. Since preferences are subjective, offering diverse plans covering all possible combinations of those factors would benefit the astronaut to pick the plans that fit their needs. For example, choosing a plan that goes through a route with more sun exposure to charge the rovers’ batteries during sample collection, which was not considered in the model in the first place.
Another usage for behaviour planning could be used in multi-objective planning~(\cite{geisser2022admissible}), which seeks to find a Pareto optimal set of plans for a set of objectives. In this case, the objectives could be the number of rovers used in the planning problem.

\section{Conclusions \& Future work}\label{sec:conclusion-future}

Diverse planning is a method for generating multiple distinct plans. It is useful in real-world applications when a single plan is not sufficient to account for, for example, user preferences, different future situations, or practical execution challenges.
This paper addresses the current challenges in diversity planning, particularly the limitations of representing diversity as a distance function. It introduces a new formulation that suggests a qualitative approach to diversity representation based on an n-dimensional grid called the behaviour space, where each dimension represents a feature. Furthermore, it introduces a diverse planning framework called behaviour planning, which incorporates the diversity model during the planning phase, enabling the generation of more diverse plans compared to current diverse planning approaches.
Behaviour planning comprises two components: a model-based approach that allows for detailed diversity representation and directs the planner to generate plans with different orderings. We demonstrated one implementation of behaviour planning using planning-as-satisfiability. The expressivity offered by planning-as-satisfiability enables support for various planning tasks, including over-subscription and numerical planning.
To evaluate the practicality of the behaviour planning framework in generating diverse plans, we conducted experiments. Our results showed that our implementation is valid for generating diverse plans compared to state-of-the-art diverse planners for classical planning. Additionally, two further experiments demonstrated that behaviour planning is the first diverse approach supporting over-subscription and numerical planning. We demonstrated when to use the new diversity planning formulation, how it works, and the benefits users can gain from using behaviour planning.

Behaviour count and \texttt{MaxSum} are two metrics used to quantify the diversity of a set of plans. While they both serve this purpose, they have distinct advantages and drawbacks. For instance, behaviour count provides information about the number of different plans included in a set of plans, based on a customisable diversity model. In contrast, \texttt{MaxSum} cannot provide this information. However, behaviour count also has limitations. Consider two sets of plans with the same behaviour count. In this case, \texttt{MaxSum} can distinguish between them, while behaviour count cannot. In case both sets have the same \texttt{MaxSum}, then the user would need to manually check the behaviours per each set and pick the set that satisfies their needs. A potential future work involves distinguishing between two sets of behaviours based on their level of dissimilarity (i.e., similar and dissimilar features values). Another research direction is investigating guided search strategies instead of the current blind search approach used by $\texttt{FBI}_\texttt{X}$. Filtering the behaviour space by removing invalid behaviours would improve $\texttt{FBI}_\texttt{X}$ coverage. Therefore, another research direction is to explore techniques to reduce the dimensionality of the search space to expedite the search process. Finally, expanding the library of behavioural features is another possible research venue.

\bibliographystyle{unsrtnat}
\bibliography{references}  






\end{document}